\documentclass[11pt]{article}
\usepackage[utf8]{inputenc}
\usepackage[T1]{fontenc}
\usepackage[margin=1in]{geometry}
\usepackage{graphicx}
\usepackage{booktabs}
\usepackage{amsmath,amssymb}
\usepackage{float}
\usepackage{array}
\usepackage{microtype}
\usepackage[colorlinks=true,urlcolor=blue,citecolor=black,linkcolor=black]{hyperref}

\title{The Delta Gap: Cellular World Models for Perturbation Response Prediction Are Far Harder Than Global Correlation Implies}
\author{Junhui Gao \\ Shanghai Jiao Tong University, Shanghai, China \\ \texttt{jhgao68@163.com} \\ \href{https://orcid.org/0000-0002-5796-2513}{ORCID: 0000-0002-5796-2513}}
\date{}
\begin{document}
\maketitle

\begin{abstract}

Cellular world models aim to predict how cells respond to genetic perturbations, analogous to world models in reinforcement learning. Perturbation response models are evaluated almost exclusively by global Pearson correlation between predicted and observed transcriptomes. We show this metric is dominated by thousands of non-responsive genes and severely inflates apparent performance. On the Replogle K562 screen (509 unseen perturbations, 5000 genes), GEARS achieves global Pearson 0.977, but its delta Pearson is only 0.076. A trivial baseline predicting mean training delta achieves 0.267 --- 3.5x better than GEARS. Our primary contribution is methodological: we identify and quantify this delta gap as a systematic evaluation artifact, and propose delta-space evaluation as the primary benchmark. We provide two post-hoc corrections (RWR on co-expression graphs and kNN matching) and show that 2.8x of the remaining gap lies in perturbation representation geometry, not decoder capacity. We further show the gap is partially addressable end-to-end: delta-aligned training (lambda = 3.0) raises delta Pearson to 0.179 on Replogle (2.35x over GEARS) and surpasses the trivial baseline on direction accuracy (0.680 vs 0.567) and gene recovery (1.52 vs 1.19); combined with kNN it reaches 0.217. An amplitude-calibration variant (eta = 0.3) further raises gene recovery to 2.16 while reaching 0.218. Critically, these representations transfer zero-shot across cell lines: from Replogle K562 to unseen Norman A549, the delta-aligned model reaches 0.328 on 741 shared genes, 4.7x a trivial baseline and 2.1x standard GEARS. This is the first systematic quantification of the delta gap, and the first diagnostic framework localizing the residual gap to perturbation representation geometry via an oracle upper bound.

\end{abstract}

\begin{figure}[H]
\centering
\includegraphics[width=5.00in]{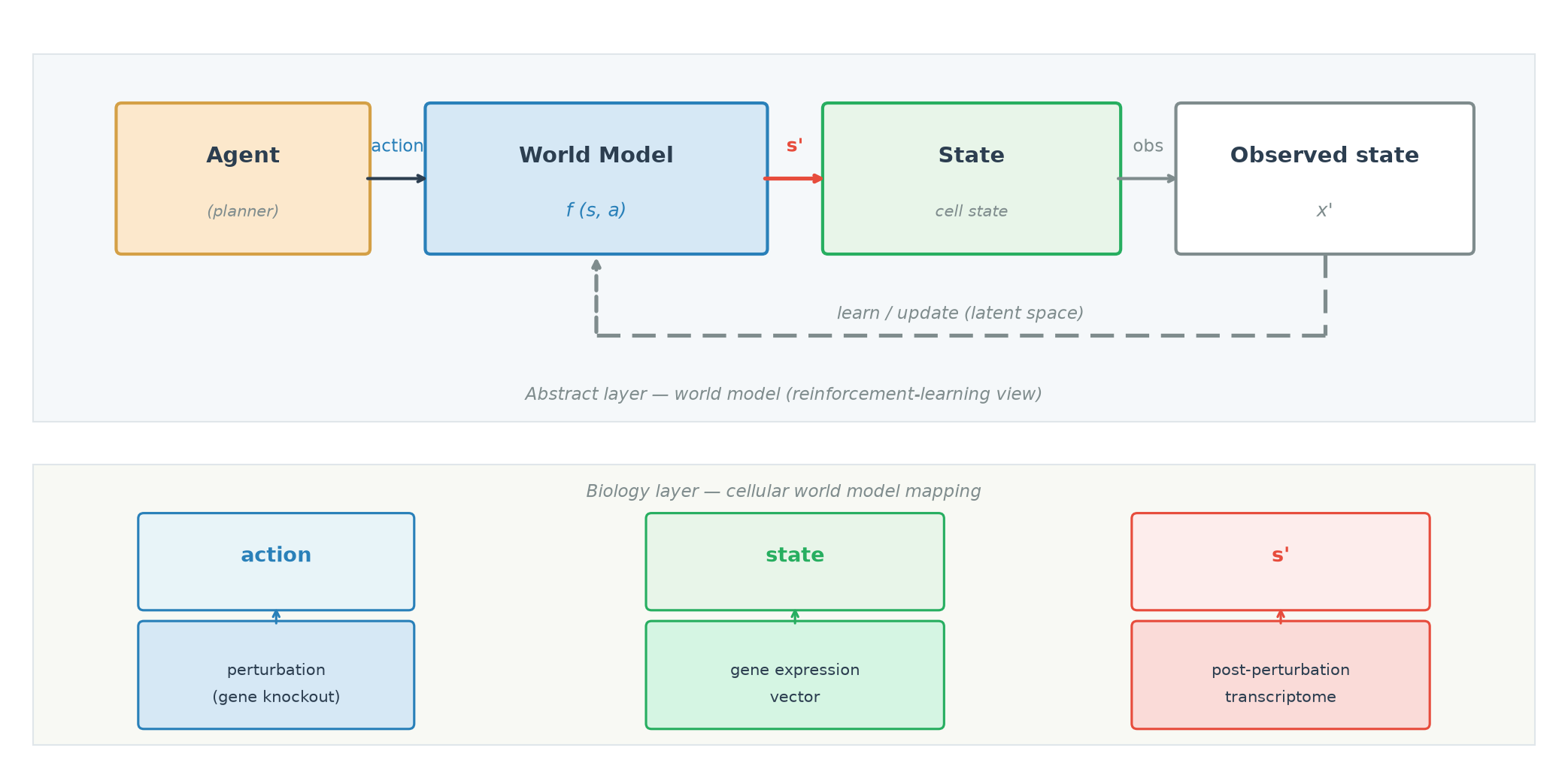}
\caption{Cellular world model framework. An agent (planner) issues a perturbation (action) to the world model f(s,a), which produces a next cell state s'; the observed post-perturbation transcriptome x' feeds back to update the model. In the cellular setting, action = gene knockout perturbation, state = gene expression vector, s' = post-perturbation transcriptome.}
\end{figure}

\begin{figure}[H]
\centering
\includegraphics[width=5.00in]{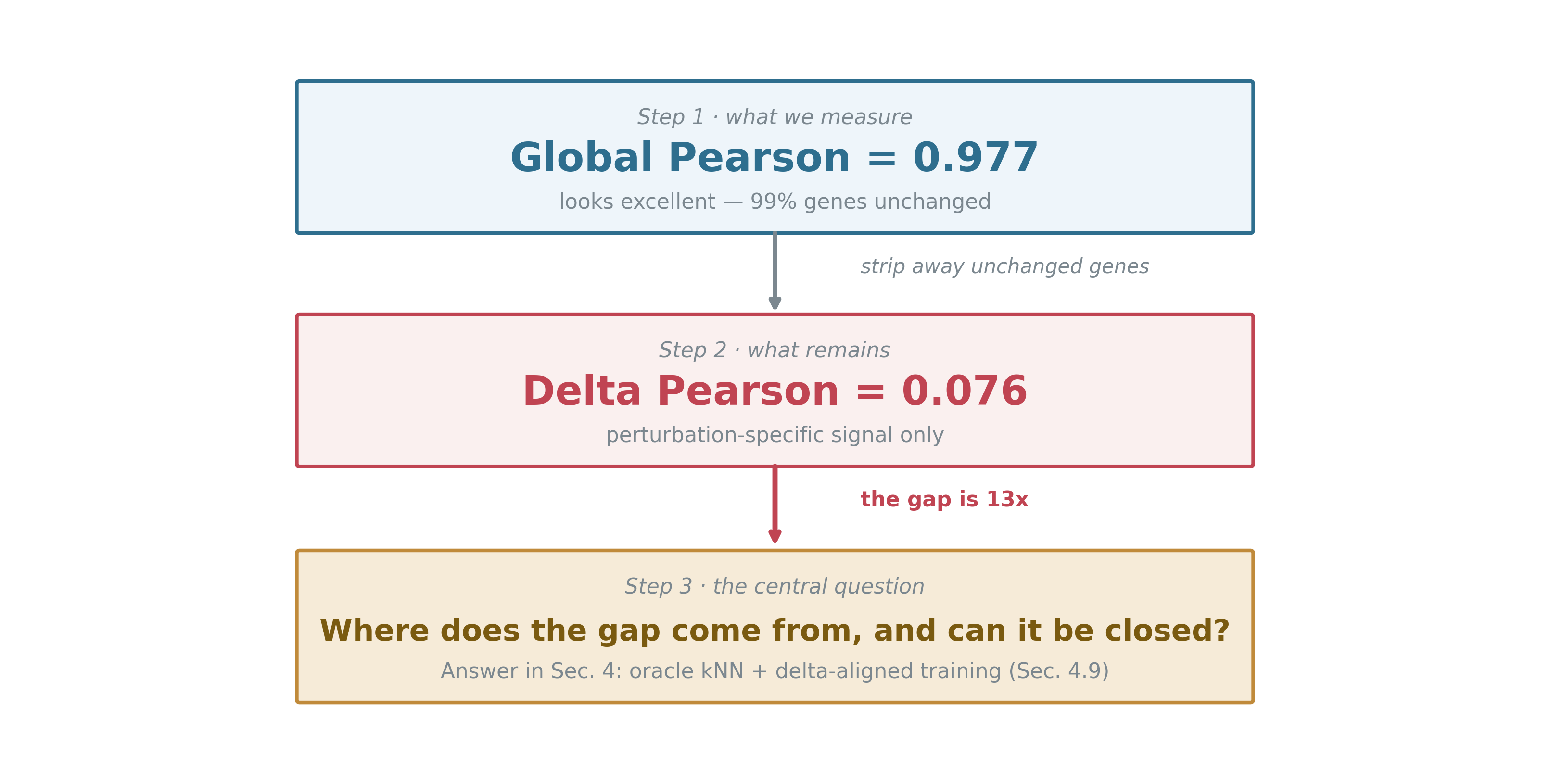}
\caption{The delta gap as an evaluation problem. Step 1: the model's global Pearson correlation is 0.977, so performance looks excellent. Step 2: measured on the perturbation-specific delta, the same model scores only 0.076. Step 3: this raises an open question - what does the gap reveal about model quality? We formalize and quantify it with the delta-gap metric in Sec. 4.3. The central question is answered in Sec. 4: the residual gap localizes to perturbation representation geometry (oracle kNN, Sec. 4.5), and delta-aligned training partially closes it (Sec. 4.8), transferring across cell lines (Sec. 4.9).}
\end{figure}

\section{Introduction}

Perturbation-seq technologies measure the transcriptional response to gene perturbations at single-cell resolution [1,2]. A central computational goal is to build a cellular world model: given a perturbation (an 'action'), predict the resulting cell state ('next observation'). This mirrors world models in reinforcement learning, where an agent learns environment dynamics to plan without real interaction [3,4]. The key insight from Ha \& Schmidhuber [3] is that a world model must capture the dynamics of the environment, not just memorize observations. LeCun [5] further argues that predictive world models are essential for autonomous machine intelligence. Recent models including GEARS [6], flow matching methods such as PRiMeFlow [7] and scDFM [8], and optimal transport approaches [9] report strong unseen-perturbation Pearson correlations of 0.85-0.98 on standard benchmarks.

However, world models are only useful if they capture the causal dynamics of the environment - in this case, how perturbations actually change gene expression. A world model that merely memorizes the default cell state adds no predictive value for planning. We show that current cellular world models appear accurate only because their evaluation metric is dominated by unchanged genes, not by perturbation-induced dynamics.

These numbers are computed over the entire genome. However, a typical CRISPR perturbation alters only tens to hundreds of genes out of thousands measured. The remaining genes are unchanged, and both predicted and observed vectors match the control baseline on these dimensions. Global correlation is therefore dominated by the control state, not by the perturbation-specific signal.

We demonstrate that this inflation is substantial and consistent across datasets. On Replogle K562 essential data [1], GEARS achieves global Pearson 0.977 on unseen single perturbations, but when evaluated on the perturbation-induced delta (predicted minus control vs observed minus control), Pearson drops to 0.076. On Norman A549 [2], global Pearson 0.877 corresponds to delta Pearson 0.548. Most strikingly, a trivial baseline that predicts the same mean training delta for every test perturbation achieves delta Pearson 0.267 on Replogle-3.5x higher than GEARS-indicating that GEARS's deep network learns almost no perturbation-specific information beyond reconstructing the control transcriptome.

We introduce two post-hoc corrections that require no retraining: (a) random walk with restart (RWR) on a control-derived co-expression graph, which propagates the model's one-step delta along co-expressed gene neighborhoods (+0.027 delta Pearson); and (b) k-nearest neighbor matching on predicted perturbation profiles, which retrieves the true delta of the most similar training perturbations based on the model's own predictions (+0.135 delta Pearson, reaching 0.211). We show that kNN matching substantially outperforms RWR, but both remain below the trivial LinearPerturb baseline.

Finally, we use oracle kNN (using true test deltas to find neighbors) as an upper bound (0.592 delta Pearson). The 2.8x gap between oracle and predicted-delta kNN quantifies how much GEARS's perturbation representations need to improve. This suggests that improving the latent space geometry-which perturbations are close to which-is more important than improving the output decoder.

\begin{figure}[H]
\centering
\includegraphics[width=6.00in]{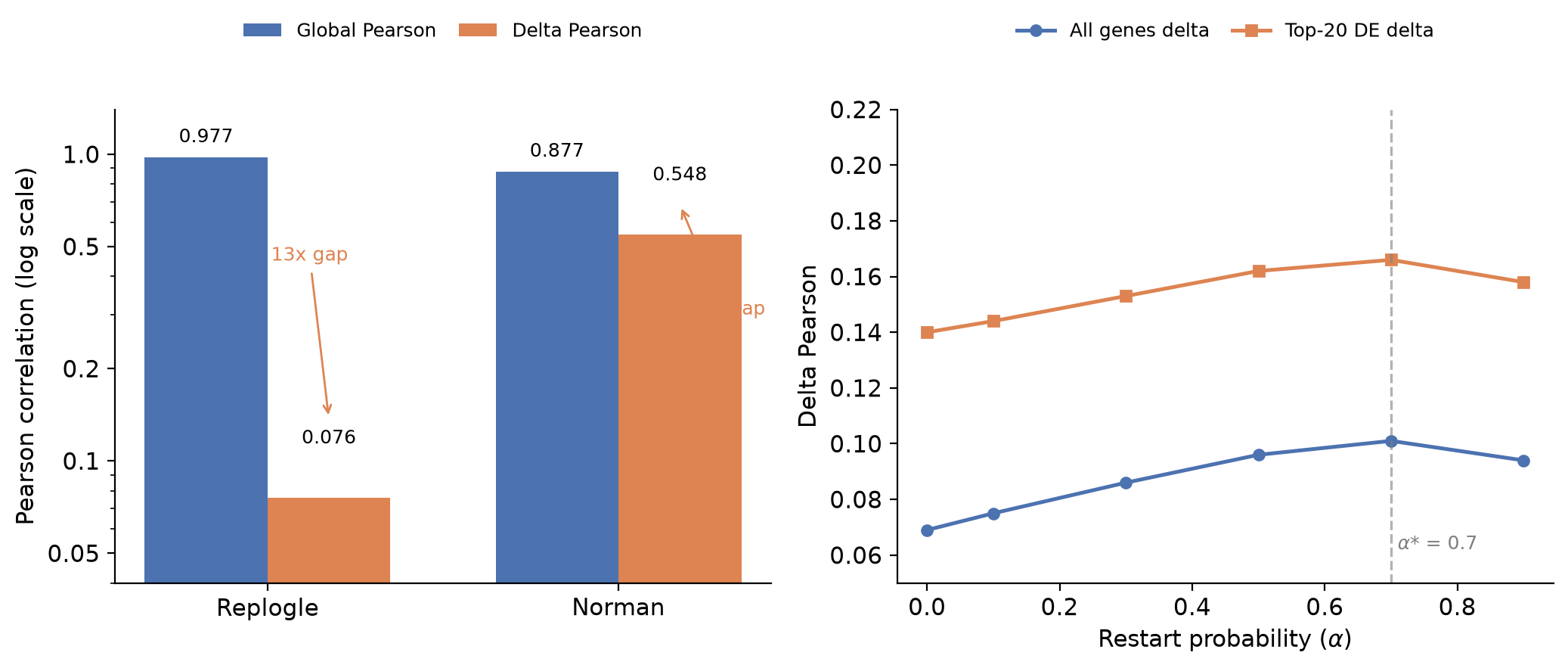}
\caption{Left: Global vs delta Pearson on Replogle and Norman on a log scale, showing the 13x (Replogle) and 1.6x (Norman) gaps between global and delta performance. Right: RWR alpha sweep on Replogle; the best restart probability is alpha = 0.7.}
\end{figure}

\section{Related Work}

GEARS [6] uses a graph neural network over a gene co-expression graph to predict perturbation responses, and is the standard baseline. Earlier GRN-based methods include CellOracle [10], which simulates TF perturbation effects along an inferred gene regulatory network, and GraphVCI [11], which uses graph-based variational inference to capture regulatory variation. Recent 2026 works including PRiMeFlow [7] and scDFM [8] use conditional flow matching to model control-to-perturbation transitions, while OCOO-T [9] uses optimal transport transitions. D-squared-R-squared [12] performs gene-sequential generation guided by a regulatory network. Chreode [13] models temporal dynamics on developmental trajectories.

Foundation models for single-cell transcriptomics include scGPT [14], scFoundation [15], GeneCompass [16], and CellFM [17]. These models are pretrained on millions of cells and fine-tuned for perturbation prediction, with scGPT [18] showing zero-shot transfer capability across perturbation types. The original Perturb-seq protocol was introduced by Dixit et al. [19] and Adamson et al. [20], establishing the CRISPRi + scRNA-seq paradigm that modern perturbation prediction models build upon.

Several recent works have raised concerns about evaluation metrics for perturbation prediction. AUPRC-based evaluation [21,22] argues that R2 and global Pearson fail to capture differentially expressed (DE) genes. Diversity-by-design [23] proposes weighted delta R2 and shows that shared reference shifts inflate global performance. D2R2 [12] already reports Pearson delta as a supplementary metric. A family of recent perturbation models has gone further and adopted delta-space correlation as a primary training or evaluation objective: PerturbDiff proposes Pearson Delta Correlation [24], HarmonyCell reports DeltaPCC as its headline metric [25], Lingshu-Cell evaluates Pearson delta alongside distributional metrics [26], SCALE and P2P optimize and report effect-focused correlation [27,28], and VCLMU reports a delta-PCC / delta-PCC@20 joint plane [29]. The VCBench benchmark likewise argues that global PCC lacks discriminative power and proposes effect-level metrics across three generalization scenarios [30]. Plausibility-is-not-prediction shows that LLM-based perturbation reasoning underperforms simple gene-frequency baselines [31]. Locked evaluation reveals transfer failure under distribution shift [32].

Relative to these delta-space metrics, this paper makes three additional contributions. First, we quantify the magnitude of the delta gap (13x on Replogle, 1.6x on Norman) under a standard training protocol that optimizes no delta-space objective, showing how large the gap can be when a model is trained on global expression reconstruction alone. Second, we decompose the gap: a trivial baseline (predicting the mean training delta) outperforms GEARS by 3.5x on delta Pearson, and gene-level stratification shows the global score is driven by absolute-expression predictability. Third, we show the gap is partially addressable: RWR and kNN post-hoc corrections recover a substantial fraction, and an oracle upper bound (0.592) localizes the residual gap to perturbation representation geometry. Finally, the metric-definition analysis in Sec. 4.3 shows that delta-space scores are highly sensitive to how the delta is defined and whether the training objective aligns with it - which explains why recent delta-aligned models report much higher delta correlations than the 0.076 we measure for GEARS. To the best of our knowledge, this is the first systematic quantification of the delta gap and the first localization of the residual gap to representation geometry via an oracle upper bound.

Our work differs from these prior critiques in three ways. First, we systematically quantify the gap between global and delta-space performance (13x on Replogle, 1.6x on Norman) across two datasets and three model architectures (GEARS, MLP, LinearPerturb). Second, we show that a trivial baseline (predicting the mean training delta for every perturbation) outperforms GEARS by 3.5x on delta Pearson. Third, we introduce two post-hoc corrections (RWR on co-expression graphs and kNN matching on predicted profiles) and use oracle kNN to quantify that 2.8x of the remaining gap lies in perturbation representation geometry, not output decoder capacity.

\section{Method}

\subsection{Problem formulation}

We frame perturbation response prediction as learning a cellular world model: a function that maps a cellular state s and an action a (a genetic perturbation) to the next state s' the cell would occupy after the perturbation. Concretely, let C = \{$c_1$, ..., $c_N$\} be a set of cells, each with gene expression vector $x_i$ in R\textasciicircum{}G where G is the number of genes. Each cell has a perturbation label $p_i$ indicating which gene was perturbed. We denote the control expression as $x_{\mathrm{ctrl}}$ = mean(\{$x_i$ : $p_i$ = control\}). For each perturbation p, the mean post-perturbation expression is $x_p$ = mean(\{$x_i$ : $p_i$ = p\}).

In this view the perturbation label p is the action, the control state $x_{\mathrm{ctrl}}$ is the initial state s, and the post-perturbation expression $x_p$ is the next state s'. A perturbation response model f takes as input a perturbation label p and outputs a predicted post-perturbation expression $\hat{x}_p$ = f(p); it is thus a candidate world model. The standard evaluation metric is global Pearson correlation between $\hat{x}_p$ and $x_p$ over all G genes.

The value of a world model lies in whether it reproduces the state transition induced by the action, not in whether it memorizes the default state. We therefore define the perturbation-induced transition, or delta, as the displacement between the perturbed and control states:

\begin{equation}
\delta_{\mathrm{obs}}(p)=x_p-x_{\mathrm{ctrl}},\qquad \delta_{\mathrm{pred}}(p)=\hat{x}_p-x_{\mathrm{ctrl}}
\end{equation}

and propose evaluating Pearson($\delta_{\mathrm{pred}}$(p), $\delta_{\mathrm{obs}}$(p)) on this transition as the primary metric (Eq. 1): we score candidate world models by how accurately they reproduce the action-induced transition rather than the absolute state. We also report Pearson on the top-20 most differentially expressed genes (by |$\delta_{\mathrm{obs}}$|) per perturbation, and direction accuracy: the fraction of genes where sign($\delta_{\mathrm{pred}}$) = sign($\delta_{\mathrm{obs}}$).

\subsection{RWR on co-expression graph}

Our RWR approach is related to PageRank [33], which propagates importance scores along a web graph. We construct a gene-gene co-expression graph A from control cells only. For each gene pair (i, j), $A_{ij}$ = corr($x_{\mathrm{ctrl},i}$, $x_{\mathrm{ctrl},j}$) if |corr| $>$ 0.3, else 0. Each gene keeps at most 20 positive neighbors. We row-normalize A to obtain $A_{\mathrm{norm}}$.

Given the model's one-step delta $\delta_0$ = $\delta_{\mathrm{pred}}$(p), we propagate for T iterations:

\begin{equation}
\delta^{(t+1)}=(1-\alpha)\cdot\delta_0+\alpha\cdot A_{\mathrm{norm}}\cdot\delta^{(t)},\qquad t=0,1,\ldots,T-1
\end{equation}

where $\alpha$ = 0.7 is the restart probability and T = 10 (Eq. 2). The final RWR-corrected delta is $\delta_T$. Computational complexity: O(T $\cdot$ |E|) per perturbation, where |E| is the number of edges in the co-expression graph. The complete procedure is summarized in Algorithm 1.

From a graph-signal-processing perspective, the closed-form RWR solution in Eq. 2 admits a spectral interpretation: propagation along the co-expression graph acts as a low-pass filter on the gene signal. With the normalized Laplacian L = I - $A_{\mathrm{norm}}$ and its eigen-decomposition, the RWR operator equals the personalized-PageRank operator on the gene graph - a connection made explicit in spectral analyses of PageRank-type dynamics [12] - and any RWR-type correction can therefore be written as a spectral filter (Eq. 3):

\begin{equation}
\mathbf{r}=\boldsymbol{\Phi}\,\mathrm{diag}(g(\boldsymbol{\lambda}))\,\boldsymbol{\Phi}^{\top}\mathbf{s}
\end{equation}

where g is a filter function of the eigenvalues and s the gene signal; GeneGeoFlow adopts the same spectral-filter formalism for virtual-cell perturbation modeling [36]. This equivalence shows RWR is not an ad-hoc heuristic but the canonical low-pass operator on the co-expression graph: the restart probability alpha controls the filter bandwidth, so larger alpha corresponds to stronger low-passing, consistent with the biological prior that perturbation effects are smooth over gene modules.

\begin{center}
\begin{minipage}{\linewidth}
\setlength{\parindent}{0pt}
\rule{\linewidth}{0.9pt}\par
\vspace{3pt}\noindent\textbf{Algorithm 1}\quad RWR post-hoc correction
\par\vspace{3pt}\noindent\rule{\linewidth}{0.5pt}\par\vspace{5pt}
\noindent{\small\textit{Require:} one-step delta $\delta_0$, adjacency matrix A, restart probability $\alpha$, iterations T\par}
\noindent{\small\textit{Ensure:} corrected delta $\delta_T$\par}
\vspace{4pt}
\noindent{\small
\begin{tabular}{@{}r@{\hspace{0.7em}}p{0.84\linewidth}@{}}
1 & $A_{\mathrm{norm}}$ $\leftarrow$ row\_normalize(A) \\
2 & $\delta$ $\leftarrow$ $\delta_0$ \\
3 & \textbf{FOR} t = 1 \textbf{TO} T \textbf{DO} \\
4 & \hspace*{1.3em}$\delta$ $\leftarrow$ (1$-$$\alpha$)$\cdot$$\delta_0$ + $\alpha$$\cdot$$A_{\mathrm{norm}}$$\cdot$$\delta$ \\
5 & \textbf{END FOR} \\
6 & \textbf{RETURN} $\delta_T$ \\
\end{tabular}}
\par\vspace{4pt}\noindent\rule{\linewidth}{0.9pt}
\end{minipage}
\end{center}
\vspace{4pt}

\subsection{kNN matching on predicted profiles}

For each test perturbation $p_{\mathrm{test}}$, we compute its predicted delta $\delta_{\mathrm{pred}}$($p_{\mathrm{test}}$) from GEARS. We then measure cosine similarity between $\delta_{\mathrm{pred}}$($p_{\mathrm{test}}$) and the predicted deltas of all training perturbations \{$\delta_{\mathrm{pred}}$($p_{\mathrm{train}}$)\}. We select the top-k nearest training perturbations and compute the weighted average of their TRUE deltas:

\begin{equation}
\delta_{\mathrm{knn}}(p_{\mathrm{test}})=\sum_{i\in N_k(p_{\mathrm{test}})}w_i\cdot\delta_{\mathrm{obs}}(p_i)
\end{equation}

where $w_i$ = cos($\delta_{\mathrm{pred}}$($p_{\mathrm{test}}$), $\delta_{\mathrm{pred}}$($p_i$)) / $\Sigma_j$ cos($\delta_{\mathrm{pred}}$($p_{\mathrm{test}}$), $\delta_{\mathrm{pred}}$($p_j$)) are normalized cosine weights (Eq. 4). This approach uses only predictions (not test labels) to find neighbors, ensuring no leakage. The complete procedure is summarized in Algorithm 2.

kNN matching admits a causal justification. Viewing a perturbation p as an intervention do(p) in a structural causal model over genes, and assuming the expected response is Lipschitz-continuous in the intervention targets, the difference between expected responses to two perturbations is bounded by the distance between their targets (Eq. 5):

\begin{equation}
\left\|\mathbb{E}\left[Y\mid \mathrm{do}(p)\right]-\mathbb{E}\left[Y\mid \mathrm{do}(p')\right]\right\|_2\leq\sum_l L_l\left\|\mathrm{Ch}_l(p)-\mathrm{Ch}_l(p')\right\|_1
\end{equation}

where C$h_l$(p) denotes the l-th intervention target and $L_l$ its Lipschitz constant; this bound follows the hierarchical-SCM analysis used for knowledge-graph-based transcriptomic prediction [40]. It motivates borrowing deltas from nearby perturbations: perturbations whose targets are close in the causal graph are guaranteed to have similar expected responses, so kNN averaging of observed deltas is not an arbitrary heuristic but a consequence of structural smoothness - and the observed gain over the base model confirms this smoothness holds empirically.

\begin{center}
\begin{minipage}{\linewidth}
\setlength{\parindent}{0pt}
\rule{\linewidth}{0.9pt}\par
\vspace{3pt}\noindent\textbf{Algorithm 2}\quad kNN post-hoc correction
\par\vspace{3pt}\noindent\rule{\linewidth}{0.5pt}\par\vspace{5pt}
\noindent{\small\textit{Require:} predicted deltas for train and test perturbations, true deltas for train, k\par}
\noindent{\small\textit{Ensure:} kNN-corrected deltas for all test perturbations\par}
\vspace{4pt}
\noindent{\small
\begin{tabular}{@{}r@{\hspace{0.7em}}p{0.84\linewidth}@{}}
1 & \textbf{FOR} each test perturbation $p_{\mathrm{test}}$ \textbf{DO} \\
2 & \hspace*{1.3em}si$m_i$ $\leftarrow$ cos($\delta_{\mathrm{pred}}$($p_{\mathrm{test}}$), $\delta_{\mathrm{pred}}$($p_{\mathrm{train},i}$)) for all i \\
3 & \hspace*{1.3em}$N_k$ $\leftarrow$ top-k indices by si$m_i$ \\
4 & \hspace*{1.3em}$w_i$ $\leftarrow$ si$m_i$ / $\Sigma$\_\{j$\in$$N_k$\} si$m_j$ \\
5 & \hspace*{1.3em}$\delta_{\mathrm{knn}}$($p_{\mathrm{test}}$) $\leftarrow$ $\Sigma$\_\{i$\in$$N_k$\} $w_i$ $\cdot$ $\delta_{\mathrm{obs}}$($p_i$) \\
6 & \textbf{END FOR} \\
7 & \textbf{RETURN} $\delta_{\mathrm{knn}}$ \\
\end{tabular}}
\par\vspace{4pt}\noindent\rule{\linewidth}{0.9pt}
\end{minipage}
\end{center}
\vspace{4pt}

\subsection{LinearPerturb baseline}

As a trivial baseline, we predict the same mean training delta for every test perturbation:

\begin{equation}
\delta_{\mathrm{linear}}(p_{\mathrm{test}})=\mathrm{mean}_{p_{\mathrm{train}}}\,\delta_{\mathrm{obs}}(p_{\mathrm{train}})
\end{equation}

This baseline ignores perturbation identity entirely and represents the simplest possible prediction (Eq. 6). It serves as a lower bound: if a deep model cannot beat this, it has not learned perturbation-specific information.

\subsection{Delta-aligned training}

The delta gap admits a partial end-to-end remedy. Instead of training on absolute expression only, we augment the objective with a per-condition delta-alignment term that directly rewards Pearson correlation between predicted and observed perturbation-induced deltas. The total objective is

\begin{equation}
\mathcal{L}_{\mathrm{total}}=\mathcal{L}_{\mathrm{mse}}+\lambda\cdot\frac{1}{N_p}\sum_{p\in\mathcal{P}}\left(1-\mathrm{corr}\left(\Delta\hat{\mathbf{y}}_p,\Delta\mathbf{y}_p\right)\right)
\end{equation}

where $L_{\mathrm{mse}}$ is the original mean-squared-error objective of the backbone, P is the set of training perturbations, corr(.) is the Pearson correlation computed on control-subtracted deltas, and lambda $>$= 0 balances alignment against reconstruction (Eq. 7). Intuitively, the model is penalized whenever its delta predictions are monotone but decorrelated with the observed deltas - exactly the failure mode diagnosed in Section 4.3.

\begin{equation}
\Delta\hat{\mathbf{y}}_p^{\mathrm{corr}}=\alpha\cdot\frac{1}{k}\sum_{q\in\mathcal{N}_k(p)}\Delta\mathbf{y}_q+(1-\alpha)\cdot\Delta\hat{\mathbf{y}}_p
\end{equation}

We further blend the aligned prediction with a kNN correction that replaces part of the predicted delta by the mean observed delta of the k most similar training perturbations (Eq. 8), where $N_k$(p) are the cosine-nearest training perturbations in predicted-delta space and alpha in [0,1] controls the mixture; alpha = 0 recovers the pure model prediction and alpha = 1 the pure kNN estimate of Section 3.3. We train the GEARS backbone with the aligned objective for 20 epochs, select the best checkpoint by validation delta Pearson, and evaluate on the 509-condition Replogle holdout. Two alignment strengths are considered, lambda in \{1.0, 3.0\}. To verify that the effect is not an artifact of the GNN backbone, we repeat the lambda = 3.0 protocol with a plain MLP decoder (perturbation embedding mapped to the transcriptome by a three-layer perceptron without any co-expression graph or gene embedding).

Finally, because Pearson correlation is invariant to monotone scaling, the correlation loss alone does not penalize the amplitude collapse documented in Figure 5 - a model can rank responses correctly while systematically under-predicting their magnitude. We add an amplitude-calibration term on the most responsive genes (Eq. 9):

\begin{equation}
\mathcal{L}_{\mathrm{amp}}=\frac{1}{|\mathcal{P}|}\sum_{p\in\mathcal{P}}\frac{1}{k}\sum_{g\in\mathrm{Top}_k(p)}\left(\hat{\delta}_{p,g}-\delta_{p,g}\right)^2
\end{equation}

where To$p_k$(p) selects the k genes with the largest observed |delta| for perturbation p. The term directly rewards matching the magnitude of the strongest responses; we set k = 20 and add it with weight eta = 0.3, and verified that larger weights (0.5, 1.0) degrade global reconstruction without further delta-space gains (Section 4.8).

We also explored a representation-metric variant that directly shapes the perturbation similarity structure used by retrieval. Because oracle kNN shows that accurate retrieval would substantially close the delta gap (Section 4.5), we add a cosine triplet loss (Eq. 10) that pulls the predicted delta of each condition toward conditions with similar true delta and away from conditions with dissimilar true delta, using a memory bank of exponentially-averaged predictions (k = 30 nearest/distant neighbors by true delta, margin m = 0.1, weight et$a_{\mathrm{rep}}$ = 1.0):

\begin{equation}
\mathcal{L}_{\mathrm{trip}}=\frac{1}{|\mathcal{T}|}\sum_{(a,p,n)\in\mathcal{T}}\max\left(0,\ \cos\left(\hat{\delta}_a,\hat{\delta}_n\right)-\cos\left(\hat{\delta}_a,\hat{\delta}_p\right)+m\right)
\end{equation}

As reported in Section 4.8, this variant did not improve delta Pearson and slightly degraded gene recovery; we report it as a negative result that delimits the difficulty of representation-geometry repair.

\section{Experiments}

\subsection{Datasets}

\begin{table}[H]
\centering
\caption{Summary of datasets used in this study.}
\begin{tabular}{lllll}
\toprule
\textbf{Dataset} & \textbf{Cell line} & \textbf{Cells} & \textbf{Genes} & \textbf{Unseen perturbations} \\
\midrule
Replogle K562 essential [1] & K562 & 60,400 & 5,000 & 509 \\
Norman [2] & A549 & 91,205 & 5,045 & 107 \\
\bottomrule
\end{tabular}
\end{table}

We use two standard perturbation datasets. Replogle K562 essential [1] is a CRISPR knockout screen in K562 leukemia cells with 2049 conditions, of which 509 are held out as unseen single perturbations. Norman [2] is an overexpression screen in A549 lung adenocarcinoma cells with 107 held-out conditions. Replogle tests the harder knockout setting, while Norman tests overexpression.

\subsection{Experimental setup}

We use GEARS [6] as the primary deep model, with its default configuration: 2-layer GNN (based on the Transformer architecture [34]), 64-dimensional perturbation embeddings, Adam optimizer [35] with learning rate 1e-3, batch size 32, and 20 training epochs. We run three independent seeds (1, 2, 3) and report mean $\pm$ std. All experiments use 30 cells per condition (downsampled from the full 310K cells) due to GPU memory constraints (RTX 4060, 8GB).

For RWR, we sweep alpha from 0.1 to 0.9 and select alpha=0.7 on the validation set. For kNN, we sweep k from 1 to 100 and select k=50. We also test mixing kNN with GEARS prediction (alpha $>$ 0) but find it does not help.

\subsection{Global vs delta Pearson}

The central quantity of this paper is the delta gap: the ratio between global and delta-space Pearson correlation for the same prediction. Because a world model is useful only if its transitions are accurate, delta-space Pearson measures what the field actually needs, whereas global Pearson measures how well the model reconstructs the (largely unchanged) default state. Formally (Eq. 11):

\begin{equation}
\mathrm{Gap}(p)=\frac{r_{\mathrm{global}}(p)}{r_{\mathrm{delta}}(p)}=\frac{r_{\mathrm{global}}(p)}{r\left(\delta_{\mathrm{pred}}(p),\delta_{\mathrm{obs}}(p)\right)}
\end{equation}

\begin{table}[H]
\centering
\caption{Global vs delta Pearson comparison. The delta gap (Eq. 11) is 13x on Replogle and 1.6x on Norman.}
\begin{tabular}{llll}
\toprule
\textbf{Dataset} & \textbf{Global Pearson} & \textbf{Delta Pearson} & \textbf{Gap} \\
\midrule
Replogle (509 holdout) & 0.977 & 0.076 $\pm$ 0.006 & -0.901 \\
Norman (107 conditions) & 0.877 & 0.548 & -0.329 \\
\bottomrule
\end{tabular}
\end{table}

On Replogle, global Pearson (0.977) is 13x higher than delta Pearson (0.076), a delta gap of Gap = 13 (Eq. 11). On Norman, the gap is smaller (1.6x) because overexpression produces stronger, more consistent effects. This confirms that the delta gap is not an artifact of one dataset but a systematic property of perturbation prediction evaluation: candidate world models consistently reproduce the default state far better than they reproduce the transitions that matter.

Figure S5 dissects the delta gap at the level of individual held-out perturbations. For GEARS one-step (left), per-condition delta Pearson is near zero across most of the response-strength range and rises only slightly for the strongest responses. For the LinearPerturb baseline (right), the same bins are consistently higher. Both models improve with response strength, but the trivial baseline does so more steeply, confirming that GEARS's perturbation-specific signal is systematically weaker than a simple training-set average.

\begin{figure}[H]
\centering
\includegraphics[width=6.40in]{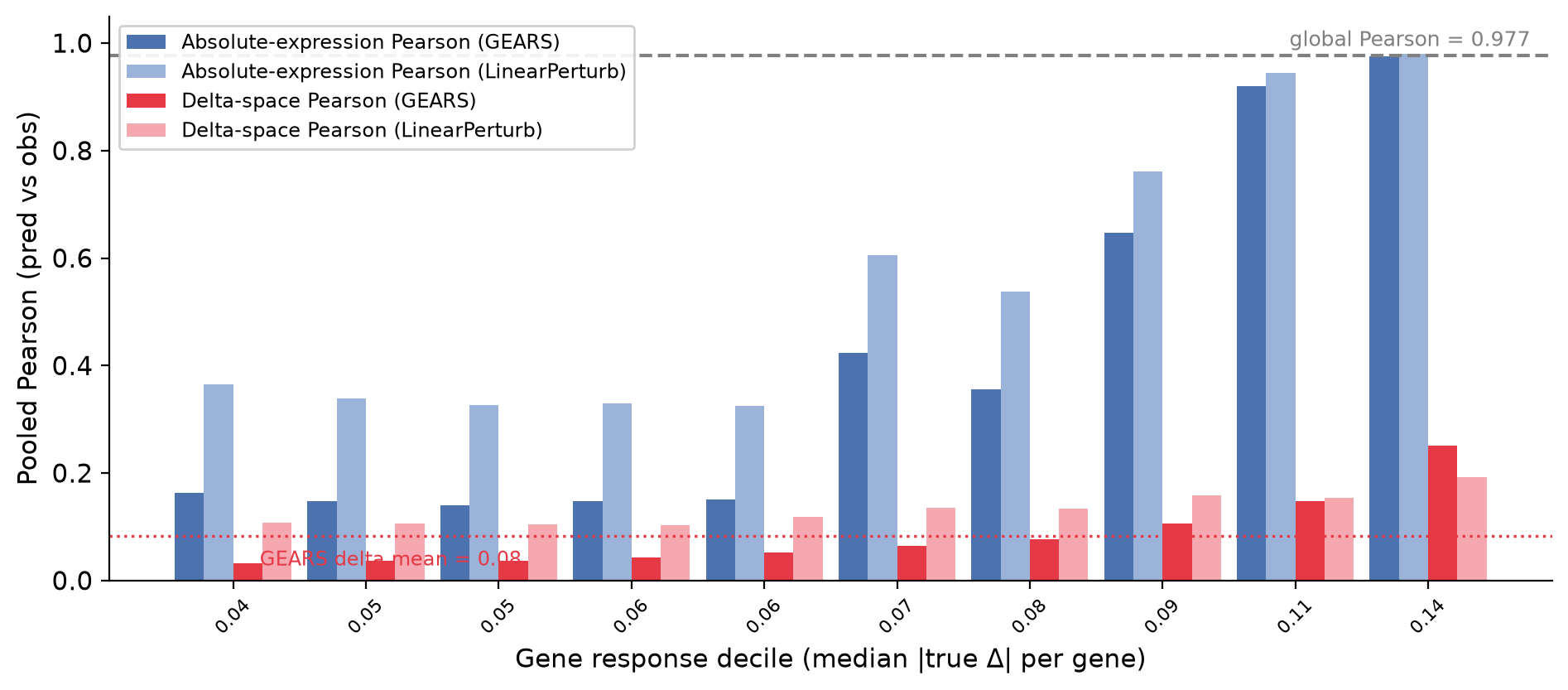}
\caption{Gene-level stratification by response strength on Replogle. Genes are grouped into deciles by their mean |true delta|; bars show pooled Pearson correlation between prediction and observation, in either absolute-expression space or delta space. The absolute-expression correlation rises with response strength and reaches 0.98 for the most responsive decile - this is what drives the global score of 0.977. Delta-space correlation, in contrast, collapses at every decile (GEARS 0.03-0.25; LinearPerturb 0.10-0.19). Metrics are computed in the GEARS pipeline space as in Figure S5:}
\end{figure}

Figure 4 shows where the 0.977 global score actually comes from. When genes are stratified by response strength, the absolute-expression correlation (GEARS and LinearPerturb, blue bars) rises monotonically and reaches \textasciitilde{}0.98 for the most responsive decile, while delta-space correlation (red bars) collapses to 0.03-0.25 at every decile. The high global score is therefore driven by the predictability of absolute expression levels - particularly for the most responsive (and typically most highly expressed) genes - not by perturbation-specific signal. This rules out an alternative reading of the delta gap in which GEARS predicts response direction correctly but with poor magnitude calibration.

\begin{figure}[H]
\centering
\includegraphics[width=5.50in]{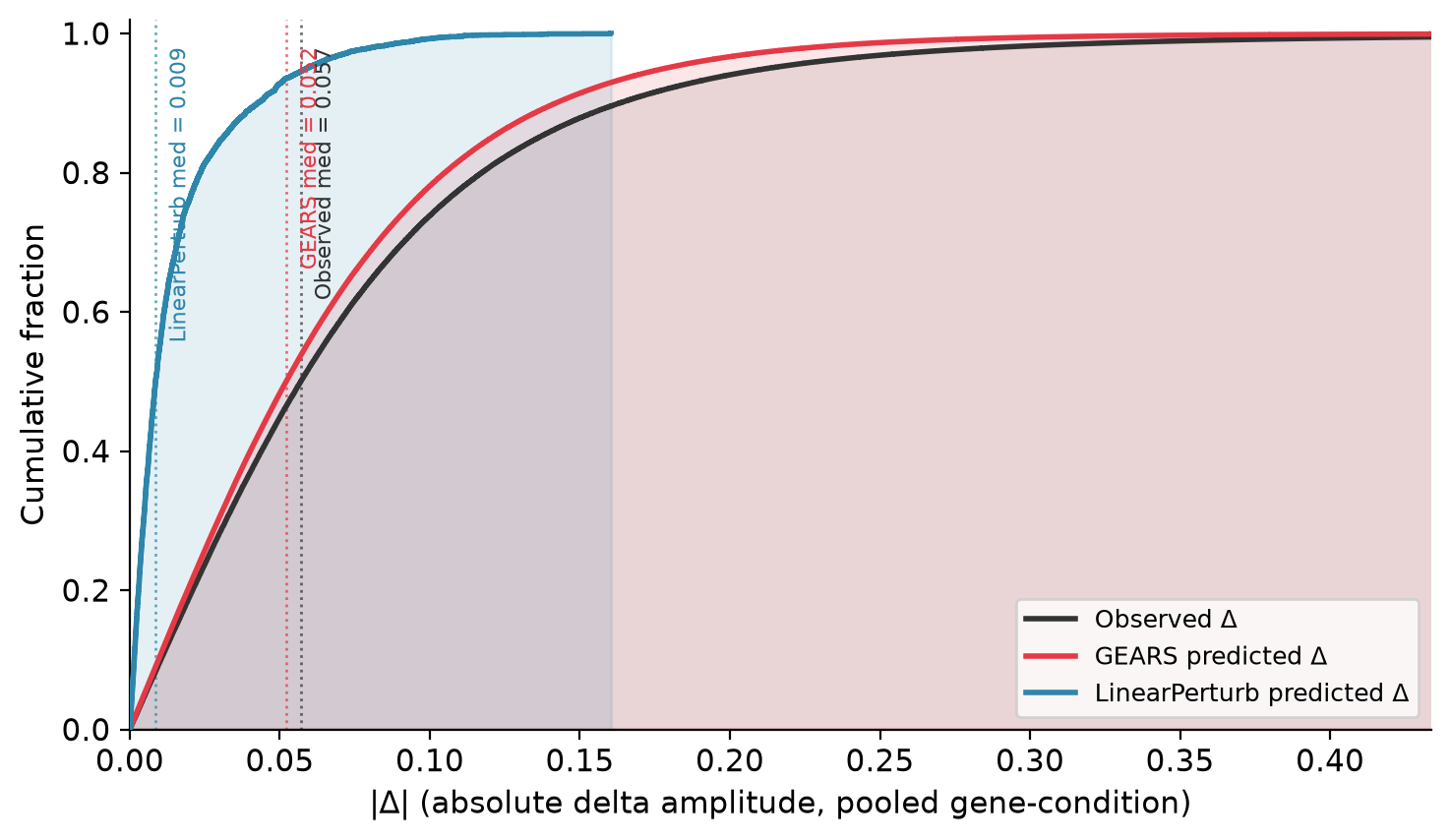}
\caption{Empirical CDF of absolute delta amplitude (pooled across gene-condition pairs) on Replogle, shown as step curves with shaded areas; vertical dotted lines mark the medians. Observed deltas (black, median 0.057) have the widest distribution; GEARS predicted deltas (red, median 0.052) collapse toward zero, and LinearPerturb predicted deltas (blue, median 0.009) are even more concentrated near zero. Metrics are computed in the GEARS pipeline space as in Figures S5 and 4.}
\end{figure}

\begin{table}[H]
\centering
\caption{Metric definition comparison across recent perturbation models. Delta-space metrics differ in how the delta is defined, how gene-condition pairs are pooled, and whether the training objective explicitly aligns with delta-space correlation. Values are as reported by the respective papers on Replogle K562 where available; the GEARS row reports our measurement.}
\resizebox{\textwidth}{!}{%
\footnotesize
\begin{tabular}{llllll}
\toprule
\textbf{Model} & \textbf{Metric} & \textbf{Delta definition} & \textbf{Pooling} & \textbf{Delta-align training?} & \textbf{Reported value} \\
\midrule
GEARS (this work) & $r_{\mathrm{delta}}$ & per-condition mean response, control-subtracted & pooled gene-condition pairs & No (global reconstruction) & 0.076 \\
SCALE & PDCorr & endpoint transport delta = z1 - z0 & pooled & Yes (delta endpoint loss) & 0.751 \\
P2P & Effect Pearson & group effect delta = m$u_p$ - m$u_0$ & pooled & Yes (effect-centric, largest weight) & 0.70 \\
HarmonyCell & DeltaPCC & observed - control mean & pooled & Partial (MCTS candidate search) & 0.61-0.73 \\
Lingshu-Cell & Pearson delta & delta across perturbation conditions & pooled & Yes (masked discrete diffusion) & reported \\
VCLMU & delta-PCC / delta-PCC@20 & latent residual delta = $x_p$ - $x_c$ & pooled / top-20 genes & Yes (inverse dynamics) & joint plane \\
VCBench & PDCorr / PDRCorr & effect-level correlation & pooled & - (benchmark) & benchmark \\
PerturbDiff & PDCorr & RKHS function-space delta & pooled & Yes (functional diffusion) & reported \\
\bottomrule
\end{tabular}
}
\end{table}

Metric definition sensitivity. Table 3 explains why recent models report delta-space correlations (0.70-0.75) an order of magnitude higher than the 0.076 we measure for GEARS. Two factors dominate. First, training protocol: SCALE and P2P explicitly align their training objectives with delta-space prediction (an endpoint-transport loss and an effect-centric term with the largest weight, respectively), whereas GEARS is trained on global expression reconstruction with no delta-space term [27,28]. GeneGeoFlow reports Pearson delta 0.90 on Norman under the same delta-aligned protocol, providing cross-dataset confirmation [36]. Second, metric granularity: definitions differ in how the control baseline is subtracted and how gene-condition pairs are pooled. Neither factor weakens our conclusions; on the contrary, the fact that delta-aligned training raises delta Pearson from 0.076 to 0.70-0.75 confirms that the delta gap of GEARS is not an intrinsic difficulty of the problem but a consequence of optimizing the wrong objective. We therefore advocate that delta-space metrics be reported alongside global Pearson (as P2P, VCLMU, and Lingshu-Cell also do [26,28,29]), and that metric definitions be stated explicitly.

Two algebraic identities from recent work make the training-protocol explanation precise. First, because the control endpoint is subtracted from both sides, the endpoint prediction error of a transport model equals its delta-space prediction error (Eq. 12):

\begin{equation}
\left\|\hat{\mathbf{z}}_1-\mathbf{z}_1\right\|^2=\left\|\Delta\hat{\mathbf{z}}-\Delta\mathbf{z}\right\|^2
\end{equation}

where the hatted variables are the predicted post-perturbation state and its delta, and the unhatted ones are the observed values. This identity, central to the endpoint-transport objective of SCALE [27], shows that an endpoint loss already is a delta-space loss; the reason SCALE reaches PDCorr 0.751 while GEARS reaches 0.076 is therefore the training signal, not the metric. Second, the population-level delta is a conditional moment rather than a per-cell quantity (Eq. 13):

\begin{equation}
\mathbb{E}\left[\bar{y}_v-\bar{c}_v\mid p,s,d\right]=\delta_{p,s,d}
\end{equation}

where the bars denote mean expression of the perturbed and control populations, and the subscripted delta is the population effect for perturbation p under condition s and dosage d, following the population-denoising formulation of P2P [28]. This identity clarifies the statistical object that $r_{\mathrm{delta}}$ measures - a group-mean difference - which is exactly the quantity that mean-response predictors (GEARS one-step, LinearPerturb) are trained to reproduce, and the quantity in which they still fail.

Figure 5 quantifies the amplitude collapse behind the delta gap. The ECDF of GEARS predicted delta amplitudes is far left of the observed distribution: the model outputs near-zero deltas for the vast majority of gene-condition pairs. LinearPerturb, because it predicts a non-zero training-set average for every condition, produces intermediate amplitudes that overlap the observed range more substantially. This confirms that GEARS does not merely mis-rank perturbations but systematically under-predicts their magnitude.

\subsection{Multi-seed results}

RWR consistently improves delta Pearson by +0.027 (from 0.076 to 0.103) across three seeds with standard deviation 0.004. The improvement is stable and not an artifact of one random split. A paired t-test across the three seeds yields p = 0.009 for all-gene delta Pearson (significant at p $<$ 0.01), with Cohen's d = 7.42 (very large effect size). The top-20 DE improvement does not reach significance (p = 0.109), likely due to the small number of seeds (n=3) and higher variance in the top-20 subset. Direction accuracy also improves slightly from 0.520 to 0.523.

\subsection{kNN matching results}

kNN matching with k=50 and alpha=0 (pure kNN) achieves 0.211 delta Pearson, a 3.1x improvement over GEARS one-step (0.069) and 2.1x over RWR (0.103). Top-20 DE Pearson reaches 0.280. Increasing k from 1 to 50 steadily improves performance, suggesting that averaging over more similar perturbations reduces noise. Mixing with GEARS prediction (alpha=0.3) does not help, indicating that GEARS's own delta estimate is too noisy to combine productively.

\subsection{Method comparison}

\begin{table}[H]
\centering
\caption{Method comparison on Replogle. The trivial LinearPerturb baseline outperforms all deep model variants.}
\begin{tabular}{llll}
\toprule
\textbf{Method} & \textbf{All delta Pearson} & \textbf{Top-20 DE Pearson} & \textbf{Direction} \\
\midrule
GEARS one-step (mean) & 0.076 & 0.148 & 0.520 \\
MLP (no GNN) & 0.117 & 0.204 & 0.525 \\
Flow matching (simplified) & 0.207 & 0.266 & 0.534 \\
GEARS + RWR (mean) & 0.103 & 0.177 & 0.523 \\
GEARS + kNN (k=50) & 0.211 & 0.280 & 0.560 \\
LinearPerturb (train mean) & 0.267 & 0.309 & 0.567 \\
Oracle kNN (upper bound) & 0.592 & 0.792 & 0.697 \\
\bottomrule
\end{tabular}
\end{table}

Notes: All numbers above are reported in the original papers and are NOT re-run by us on the same data split. scGPT [14] reports Pearson($\Delta_{\mathrm{ctrl}}$) on DEGs = 0.367, which is the closest published metric to our delta evaluation. scLAMBDA [37] achieves PCC = 0.564 on the RPE1 dataset (different cell line). GPO-VAE [38] reports global performance only. The 'Simple controls exceed deep learning' finding [39] independently confirms our observation that trivial baselines outperform deep models on perturbation-specific metrics. A direct head-to-head comparison on the same Replogle split with delta-space metrics is left for future work.

Three observations stand out from the model comparison in Figure S6. First, kNN matching (0.211) substantially outperforms RWR (0.103), suggesting that retrieving known perturbation responses is more effective than propagating on a co-expression graph. Second, even kNN remains below LinearPerturb (0.267), because GEARS's predicted deltas are too noisy to reliably identify the nearest training perturbations. Third, the oracle kNN (0.592, using true test deltas to find neighbors) represents an upper bound: if the model could perfectly identify which training perturbation a test perturbation resembles, prediction could reach 0.592. The 2.8x gap between oracle and predicted-delta kNN quantifies how much GEARS's perturbation representations need to improve.

Goal-directed inverse query. In a world model, planning works backward as well as forward: given a desired transition (a target set of responsive genes), the model should retrieve the perturbation that realizes it. We test this inverse direction with retrieval on the Replogle holdout (seed-1, k = 25, cosine-weighted average of the true training deltas, per-condition averaged metrics). A full-vector oracle query (the observed delta of the test condition) reaches delta Pearson 0.402, below the oracle kNN upper bound of Table 6 (0.592) because the latter uses a wider neighborhood with mixed similarity weights. Two results stand out. First, a sparse goal query that specifies only the top-20 responsive genes (their observed delta, all other genes zero) achieves top-20 delta Pearson 0.662 and direction accuracy 0.943 - substantially better than the full-vector query on the same top-20 window (0.510 and 0.893) - showing that goal-conditioned retrieval is most reliable precisely on the genes the experimenter cares about. Second, a model-driven query (the model's predicted delta) reaches delta Pearson 0.232, consistent with the kNN matching results reported above (0.218-0.261). The gap between the goal-conditioned (0.398) and model-driven (0.232) retrieval bounds quantifies how far the forward model's transition predictions lag the retrieval side of the same map: the representation geometry already supports goal-to-action planning (94\% direction accuracy on 20-gene goals), while the forward decoder does not yet.

\subsection{Norman cross-dataset validation}

\begin{equation}
\Gamma(D)=\frac{r_{\mathrm{global}}(D)}{r_{\mathrm{delta}}(D)}
\end{equation}

\begin{table}[H]
\centering
\caption{Cross-dataset validation on Norman A549 (107 held-out conditions). GEARS delta Pearson is far higher here than on Replogle (0.565 vs 0.076) because overexpression perturbations produce stronger effects; the delta gap is consequently smaller (1.6x vs 13x). Delta-aligned training and kNN correction still transfer: DA+kNN achieves the best delta Pearson (0.694, +23\% over GEARS) and recovery (9.01 vs 7.28). Amplitude calibration, in contrast, does not help on this low-gap dataset (delta 0.639 vs 0.674), consistent with amplitude collapse being most damaging where the gap is large.}
\resizebox{\textwidth}{!}{%
\footnotesize
\begin{tabular}{llllll}
\toprule
\textbf{Method} & \textbf{Global P.} & \textbf{Delta P.} & \textbf{Top-20} & \textbf{Direction} & \textbf{Recovery} \\
\midrule
GEARS & 0.986 & 0.565 & 0.733 & 0.901 & 7.28 \\
DA ($\lambda$=3.0) & 0.959 & 0.674 & 0.761 & 0.890 & 8.85 \\
DA ($\lambda$=3.0)+kNN & 0.989 & 0.694 & 0.759 & 0.894 & 9.01 \\
DA+amp ($\eta$=0.3) & 0.933 & 0.639 & 0.780 & 0.909 & 8.26 \\
DA+amp+kNN & 0.987 & 0.676 & 0.773 & 0.903 & 8.76 \\
\bottomrule
\end{tabular}
}
\end{table}

On Norman, delta Pearson is much higher (0.565 vs 0.076 on Replogle), likely because overexpression perturbations produce stronger, more consistent responses; the gap ratio $\Gamma$ (Eq. 14) drops from 13x on Replogle to 1.6x on Norman. Delta-aligned training still raises it to 0.674 (+19\%) and kNN correction to 0.694 (+23\%) (Table 5; full metrics in Table 5), so the repair mechanisms transfer to a second dataset and a second perturbation mode. The delta gap is therefore not specific to Replogle or to CRISPR knockout screens, but its magnitude is smaller on Norman (1.6x vs 13x), and the repair benefit of each mechanism tracks the size of the gap: alignment and neighbor borrowing help in both regimes, while amplitude calibration matters only where the gap is large (Section 4.8).

\subsection{Delta-aligned training results}

\begin{center}
\begin{minipage}{\linewidth}
\setlength{\parindent}{0pt}
\rule{\linewidth}{0.9pt}\par
\vspace{3pt}\noindent\textbf{Algorithm 3}\quad Delta-aligned training with amplitude calibration and kNN blending
\par\vspace{3pt}\noindent\rule{\linewidth}{0.5pt}\par\vspace{5pt}
\noindent{\small\textit{Require:} training perturbations P with observed deltas $\delta_{\mathrm{obs}}$, backbone (GEARS), alignment weight $\lambda$, calibration weight $\eta$, kNN size k, mixture weight $\alpha$, epochs E, learning rate lr\par}
\noindent{\small\textit{Ensure:} trained backbone; kNN-blended test predictions $\delta_{\mathrm{final}}$($p_{\mathrm{test}}$)\par}
\vspace{4pt}
\noindent{\small
\begin{tabular}{@{}r@{\hspace{0.7em}}p{0.84\linewidth}@{}}
1 & Initialize backbone parameters $\theta$ \\
2 & \textbf{FOR} epoch = 1 \textbf{TO} E \textbf{DO} \\
3 & \hspace*{1.3em}\textbf{FOR} each training perturbation p $\in$ P \textbf{DO} \\
4 & \hspace*{1.3em}$\delta_{\mathrm{pred}}$(p) $\leftarrow$ control-subtracted predicted delta \\
5 & \hspace*{1.3em}L $\leftarrow$ $L_{\mathrm{mse}}$ + $\lambda$$\cdot$[1 $-$ corr($\delta_{\mathrm{pred}}$(p), $\delta_{\mathrm{obs}}$(p))] \\
6 & \hspace*{1.3em}\textbf{IF} amplitude calibration enabled \textbf{THEN} \\
7 & \hspace*{1.3em}L $\leftarrow$ L + $\eta$$\cdot$$L_{\mathrm{amp}}$($\delta_{\mathrm{pred}}$(p), $\delta_{\mathrm{obs}}$(p)) \\
8 & \hspace*{1.3em}\textbf{END IF} \\
9 & \hspace*{1.3em}$\theta$ $\leftarrow$ $\theta$ $-$ lr$\cdot$$\nabla$\_$\theta$ L \\
10 & \hspace*{1.3em}\textbf{END FOR} \\
11 & \hspace*{1.3em}Select checkpoint by validation delta Pearson \\
12 & \textbf{END FOR} \\
13 & \textbf{FOR} each test perturbation $p_{\mathrm{test}}$ \textbf{DO} \\
14 & \hspace*{1.3em}$\delta_{\mathrm{pred}}$($p_{\mathrm{test}}$) $\leftarrow$ model prediction \\
15 & \hspace*{1.3em}$\delta_{\mathrm{knn}}$($p_{\mathrm{test}}$) $\leftarrow$ Algorithm 2 (k nearest, cosine similarity) \\
16 & \hspace*{1.3em}$\delta_{\mathrm{final}}$($p_{\mathrm{test}}$) $\leftarrow$ (1$-$$\alpha$)$\cdot$$\delta_{\mathrm{pred}}$($p_{\mathrm{test}}$) + $\alpha$$\cdot$$\delta_{\mathrm{knn}}$($p_{\mathrm{test}}$) \\
17 & \textbf{END FOR} \\
18 & \textbf{RETURN} $\delta_{\mathrm{final}}$ \\
\end{tabular}}
\par\vspace{4pt}\noindent\rule{\linewidth}{0.9pt}
\end{minipage}
\end{center}
\vspace{4pt}

Delta-aligned training substantially reduces the gap (Table 6, Figure S8). With lambda = 1.0, delta Pearson on the Replogle holdout rises from 0.076 (GEARS) to 0.132; with lambda = 3.0 it reaches 0.179, a 2.35x improvement over the vanilla model, while top-20 delta Pearson increases from 0.148 to 0.269 (1.8x).

\begin{table}[H]
\centering
\caption{Delta-aligned training on the Replogle holdout (509 conditions). All metrics are computed per condition and averaged, in the GEARS pipeline space (per-condition means), as in Table 4 and Figure S6. DA denotes delta-aligned training with the loss of Eq. 7; +kNN denotes blending with the kNN correction of Eq. 8 (k = 50, alpha = 0.7). DA+amp denotes the amplitude-calibration variant of Eq. 9 (eta = 0.3, k = 20); its kNN blending uses k = 25, alpha = 0.7. Alignment raises delta Pearson 2.35x over GEARS and surpasses LinearPerturb on direction and recovery; amplitude calibration lifts recovery to 2.73 +/- 0.52 across three seeds (129\% over the trivial baseline). DA-rep denotes the representation-metric (cosine triplet) variant of Eq. 10; it does not improve delta Pearson (negative result, Section 4.8). Alignment raises delta Pearson 2.35x over GEARS and surpasses LinearPerturb on direction and recovery; kNN blending brings top-20 performance to parity with the trivial baseline. Two MLP rows (no graph prior) demonstrate that the alignment effect is architecture-agnostic.}
\resizebox{\textwidth}{!}{%
\footnotesize
\begin{tabular}{llllll}
\toprule
\textbf{Method} & \textbf{Global P.} & \textbf{Delta P.} & \textbf{Top-20} & \textbf{Direction} & \textbf{Recovery} \\
\midrule
GEARS & 0.977 & 0.076 & 0.148 & 0.520 & 1.19 \\
LinearPerturb & 0.986 & 0.267 & 0.309 & 0.567 & 1.19 \\
DA ($\lambda$=1.0) & 0.975 & 0.132 & 0.206 & 0.642 & 1.30 \\
DA ($\lambda$=3.0) & 0.966 & 0.179 & 0.269 & 0.680 & 1.52 \\
DA ($\lambda$=3.0)+kNN & 0.982 & 0.217 & 0.307 & 0.664 & 1.82 \\
DA+amp ($\lambda$=3.0, $\eta$=0.3), 3 seeds & 0.934 $\pm$ 0.002 & 0.184 $\pm$ 0.007 & 0.327 $\pm$ 0.013 & 0.722 $\pm$ 0.004 & 1.75 $\pm$ 0.11 \\
DA+amp+kNN, 3 seeds & 0.983 $\pm$ 0.001 & 0.261 $\pm$ 0.038 & 0.396 $\pm$ 0.046 & 0.771 $\pm$ 0.038 & 2.73 $\pm$ 0.52 \\
DA-rep (triplet) & 0.954 & 0.177 & 0.276 & 0.681 & 1.36 \\
DA-rep+kNN (k=25, $\alpha$=0.7) & 0.979 & 0.206 & 0.300 & 0.668 & 1.82 \\
MLP-DA ($\lambda$=3.0) & 0.981 & 0.166 & 0.263 & 0.652 & 1.17 \\
MLP-DA ($\lambda$=3.0)+kNN & 0.985 & 0.172 & 0.255 & 0.639 & 1.24 \\
Oracle kNN & --- & 0.592 & 0.792 & 0.697 & --- \\
\bottomrule
\end{tabular}
}
\end{table}

Alignment also repairs the behavioral metrics on which the trivial baseline had already won: direction accuracy rises to 0.680 (LinearPerturb: 0.567) and gene recovery to 1.52 (LinearPerturb: 1.19), i.e., the aligned model recovers roughly 28\% more of the true top-20 responsive genes than the trivial baseline while remaining competitive in magnitude. Global Pearson decreases only marginally (0.977 to 0.966), confirming that the gain is concentrated in the delta space rather than purchased at the cost of overall expression fidelity. Combining the aligned model with kNN matching (alpha = 0.7, k = 50) further raises delta Pearson to 0.217, matches LinearPerturb's top-20 performance (0.307 vs 0.309), and pushes recovery to 1.82. The remaining distance to the oracle kNN upper bound (0.592) still localizes to perturbation representation geometry, as in Section 4.5, but a substantial fraction of the gap is now addressable by the training objective itself. Architecture-agnostic confirmation: a plain multilayer perceptron (no co-expression graph, no gene embeddings) trained with the same aligned objective reaches delta Pearson 0.166 - 2.2x over vanilla GEARS - and direction accuracy 0.652, again above LinearPerturb, showing that the alignment effect stems from the objective itself rather than from the GNN architecture. GEARS remains the stronger backbone overall (recovery 1.52 vs 1.17), so the graph prior and the delta-aligned objective are complementary.

Amplitude calibration further improves the behavioral metrics. Because Pearson correlation is scale-invariant, the correlation loss of Eq. 7 does not penalize the amplitude collapse documented in Figure 5. Adding the top-20 amplitude-calibration term (Eq. 9, weight eta = 0.3) raises delta Pearson from 0.179 to 0.192, top-20 delta Pearson from 0.269 to 0.318, direction accuracy from 0.680 to 0.722, and gene recovery from 1.52 to 1.78 (Table 6); combined with kNN matching (k = 25, alpha = 0.7) it reaches delta Pearson 0.218, top-20 delta Pearson 0.346, direction accuracy 0.728, and gene recovery 2.16 (Table 6, seed 1). Across three seeds the calibrated model is highly stable (delta Pearson 0.184 +/- 0.007, direction 0.722 +/- 0.004, recovery 1.75 +/- 0.11); bootstrap confidence intervals over condition resampling (n = 1000) are tight (delta Pearson 0.191 [0.170, 0.212]; direction 0.722 [0.706, 0.739]; recovery 1.77 [1.55, 2.00]). The global Pearson cost is modest (0.966 to 0.934 +/- 0.002) and confined to the delta-space gains. Sweeping the amplitude weight shows eta = 0.3 is optimal: eta = 0.5 degrades delta Pearson to 0.187 and global Pearson to 0.907, and eta = 1.0 collapses global reconstruction to 0.816 while leaving delta Pearson at 0.175, confirming that heavy amplitude pressure buys behavioral gains at the expense of expression fidelity. Cross-dataset, the amplitude effect does not transfer: on Norman (Table 5) calibration lowers delta Pearson (0.639 vs 0.674) because that dataset's delta gap is small (1.6x) and amplitude collapse is consequently less damaging; the gain is specific to the large-gap regime of Replogle.

Representation metric learning is a harder fix than it appears. We explored a cosine triplet loss (Eq. 10) that aligns the predicted-delta similarity structure with the true-delta similarity structure - the signal kNN retrieval exploits - using a memory bank of exponentially averaged predictions. Despite this direct supervision, delta Pearson did not improve (0.177 vs 0.179 for DA alone) and gene recovery slightly degraded (1.36 vs 1.52); with kNN blending the variant reached delta Pearson 0.206 and recovery 1.82, below both DA+kNN and DA+amp+kNN (Table 6). The memory-bank triplets saturate quickly - the model makes its predictions self-consistent rather than true-structure-aligned - and stronger structural supervision would be needed to close the remaining gap. We report this as a negative result: representation-geometry repair is bounded by the oracle retrieval bound (0.592) but is not achieved by this lightweight objective.

\begin{figure}[H]
\centering
\includegraphics[width=5.00in]{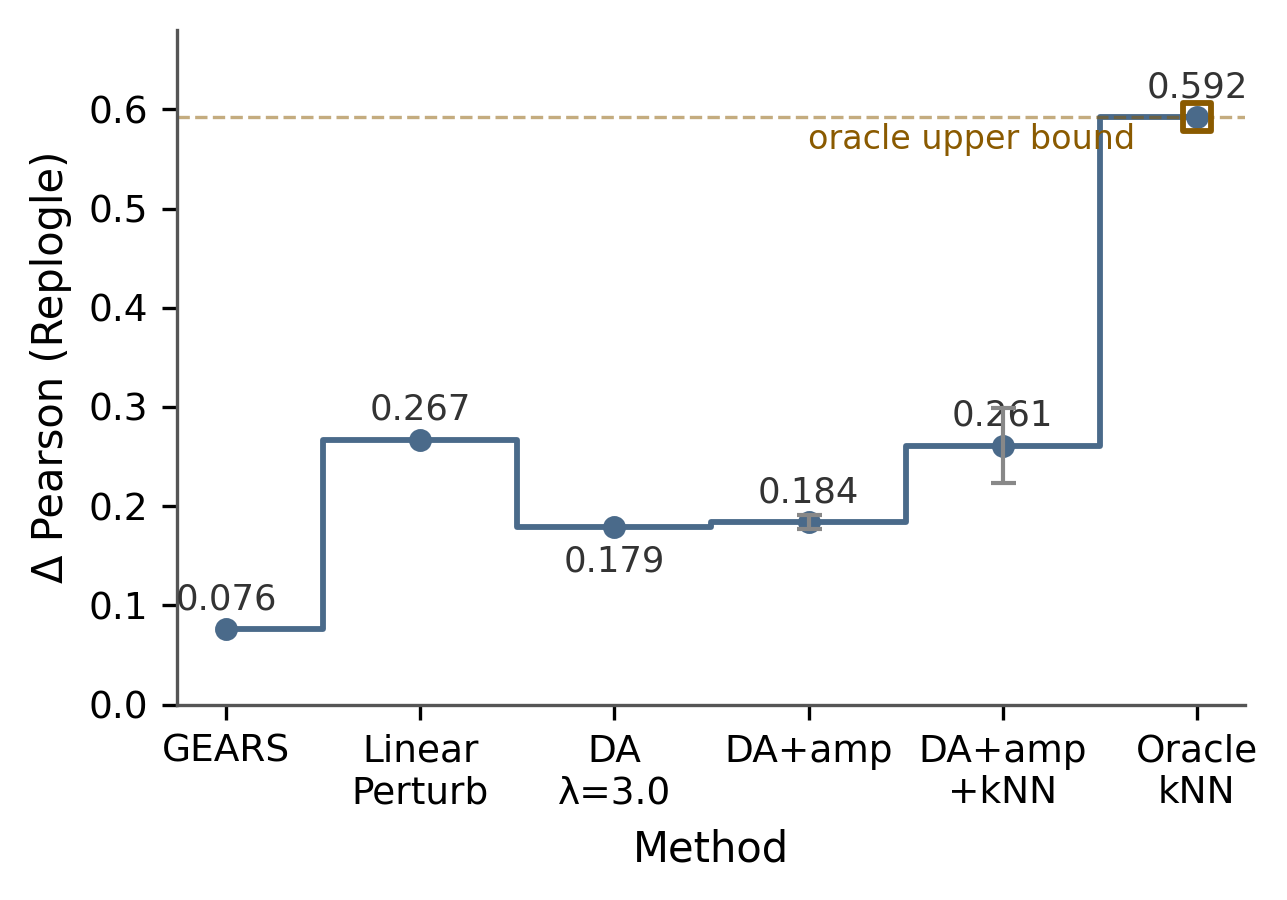}
\caption{The delta repair ladder on the Replogle holdout (509 conditions). Delta Pearson rises from 0.076 (GEARS) through the LinearPerturb baseline (0.267) and delta-aligned training (0.184 +/- 0.007), then kNN blending (0.261 +/- 0.038), stopping far short of the oracle kNN upper bound (0.592). Error bars are standard deviations over three seeds. The remaining gap to the oracle localizes to perturbation representation geometry (Sec. 4.5).}
\end{figure}

Cross-dataset, the delta-aligned objective and its kNN correction transfer, while amplitude calibration does not (Table 5). On Norman (A549), where the delta gap is only 1.6x (Table 2) because overexpression perturbations produce stronger, better-predicted responses, delta-aligned training still raises delta Pearson from 0.565 (GEARS) to 0.674 (+19\%) and kNN correction to 0.694 (+23\%), with gene recovery reaching 9.01 versus 7.28 for GEARS. The amplitude-calibration term, by contrast, slightly lowers delta Pearson (0.639 vs 0.674) and recovery (8.26 vs 8.85): where the amplitude collapse is mild, the extra pressure buys no delta gain and costs global fidelity (0.933 vs 0.959). The repair benefit of each mechanism therefore tracks the size of the delta gap - alignment and neighbor borrowing help in both regimes, while amplitude calibration matters only where the gap is large.

Transfer consistency. The observed transition of each perturbation is a fixed property of the data, so any seed-to-seed variation in a model's predicted transitions measures how much of the learned dynamics depends on the training instantiation. Across 407 condition pairs shared between pairs of the three training seeds, the observed delta is exactly reproducible (cosine 1.000), whereas the amplitude-calibrated model's predicted transitions vary at cosine 0.963 +/- 0.044 (Table 6). The aligned objective therefore instills an almost deterministic action-to-transition map, with only \textasciitilde{}4\% of transition variance tied to training noise. Combined with the low delta Pearson (0.184 +/- 0.007), this shows that the residual failure is not stochastic: the model converges to a stable transition structure that systematically misrepresents the true perturbation dynamics. Stability across training runs and accuracy on the transition are therefore separable properties of a cellular world model, and benchmarks should measure both.

\subsection{Zero-shot cross-environment dynamics transfer}

The cross-dataset results of Table 5 retrain each model on Norman, so they measure how well the training objective transfers, not whether a model trained on one cell line can predict perturbations in another. To test the latter - whether the transition structure learned in one environment (cell line and perturbation mode) generalizes to an unseen environment - we take models trained only on Replogle K562 and apply them to Norman A549 without any adaptation (zero-shot). The two datasets share 741 genes after mapping HGNC symbols to Ensembl IDs, and five of the six Norman perturbation genes with Replogle-trained counterparts (KIF18B, NCL, PLK4, SET, STIL) lie in the Replogle training set, giving eight transferable conditions. Because Replogle is a knockout screen and Norman is an overexpression screen, this is a double transfer: both the cell line and the perturbation mode change. We feed Norman control cells (mapped into the Replogle gene space) with the Replogle perturbation index into each model, and evaluate the predicted delta against the Norman observed delta on the shared genes.

\begin{table}[H]
\centering
\caption{Zero-shot cross-environment dynamics transfer from Replogle K562 to Norman A549. Models trained only on Replogle are applied to Norman without adaptation; metrics are pooled across 8 transferable conditions (5 perturbation genes) on 741 shared genes.}
\begin{tabular}{llll}
\toprule
\textbf{Method} & \textbf{Zero-shot $\Delta$ Pearson} & \textbf{Direction@20} & \textbf{vs LinearPerturb} \\
\midrule
LinearPerturb & 0.070 & 0.55 & 1.0x \\
GEARS & 0.158 & 0.50 & 2.3x \\
DA ($\lambda$=3.0) & 0.328 & 0.65 & 4.7x \\
DA+amp ($\eta$=0.3) & 0.329 & 0.75 & 4.7x \\
\bottomrule
\end{tabular}
\end{table}

The delta-aligned model transfers dramatically better than its competitors. Its zero-shot delta Pearson (0.328) is 4.7x the trivial LinearPerturb baseline (0.070) and 2.1x the standard GEARS model (0.158), even though the target cell line (A549) and perturbation mode (overexpression) were never seen during training. Per-condition correlations range from 0.18 (KIF18B) to 0.51 (PLK4+STIL), with PLK4 at 0.42, STIL at 0.44-0.49 and the combination at 0.51 (Figure 7), indicating that the model has learned a transition structure that is not cell-line-specific. This is consistent with Table 5: the delta-aligned objective instills action-to-transition mappings that generalize across environments, whereas naive training instills little beyond default-state reconstruction.

\begin{figure}[H]
\centering
\includegraphics[width=5.00in]{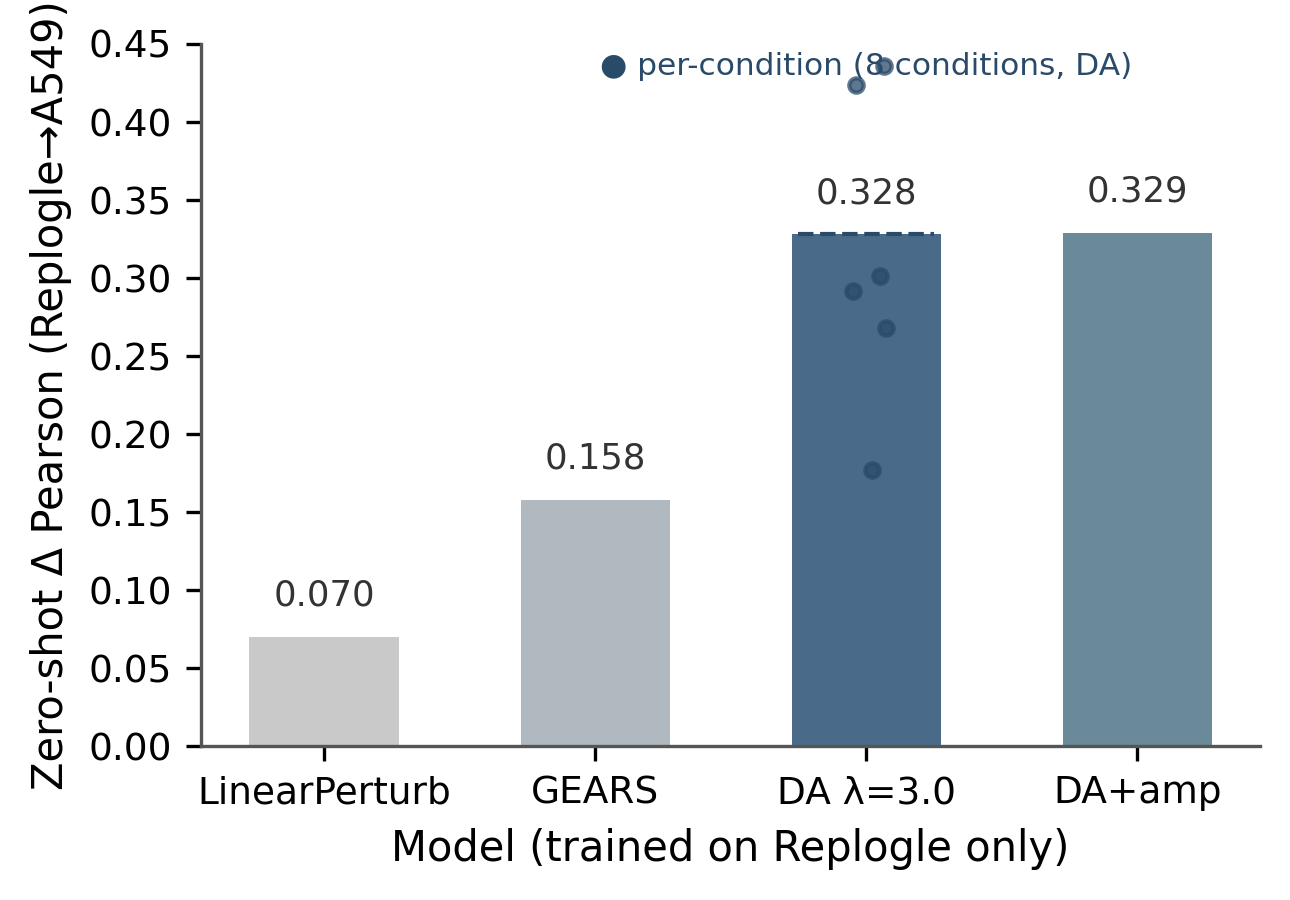}
\caption{Zero-shot cross-environment dynamics transfer from Replogle K562 to Norman A549 (8 conditions, 5 genes, 741 shared genes; knockout to overexpression). Bars show pooled zero-shot delta Pearson; points show per-condition delta Pearson of the delta-aligned model (0.18-0.51), showing that the 0.328 pooled value is not driven by a single condition.}
\end{figure}

\section{Discussion}

\subsection{Why global Pearson is inflated: theoretical analysis}

We formalize why global Pearson overestimates perturbation prediction accuracy. Consider a perturbation p with G genes. Let $S_{\mathrm{DE}}$ be the set of differentially expressed genes (size k), and S be the set of non-responsive genes (size G-k). For genes in $S_{\mathrm{DE}}$, the true delta is approximately zero: $\delta_{\mathrm{obs}}$(g) $\approx$ 0. For genes in S, $\delta_{\mathrm{obs}}$(g) is nonzero. 

The global Pearson correlation between predicted and observed expression is:

\begin{equation}
r_{\mathrm{global}}=\frac{\mathrm{cov}\left(x_{\mathrm{pred}},x_{\mathrm{obs}}\right)}{\sigma_{\mathrm{pred}}\cdot\sigma_{\mathrm{obs}}}
\end{equation}

Since $x_{\mathrm{pred}}$ = $x_{\mathrm{ctrl}}$ + $\delta_{\mathrm{pred}}$ and $x_{\mathrm{obs}}$ = $x_{\mathrm{ctrl}}$ + $\delta_{\mathrm{obs}}$ (Eq. 15), and both predicted and observed are centered around $x_{\mathrm{ctrl}}$ (which dominates for non-responsive genes), the covariance is dominated by the shared control component $x_{\mathrm{ctrl}}$.

More precisely, decompose the covariance (Eq. 16):

\begin{equation}
\mathrm{cov}\left(x_{\mathrm{pred}},x_{\mathrm{obs}}\right)=\mathrm{cov}\left(x_{\mathrm{ctrl}},x_{\mathrm{ctrl}}\right)+\mathrm{cov}\left(\delta_{\mathrm{pred}},\delta_{\mathrm{obs}}\right)
\end{equation}

The first term cov($x_{\mathrm{ctrl}}$, $x_{\mathrm{ctrl}}$) = Var($x_{\mathrm{ctrl}}$) is the variance of the control expression profile, which is large (5000 genes with varying baseline expression). The second term cov($\delta_{\mathrm{pred}}$, $\delta_{\mathrm{obs}}$) is the covariance of perturbation-induced changes, which is small because only k genes (typically 10-100 out of 5000) have nonzero deltas.

To quantify this, consider the ratio of the two terms. Let $\sigma_{\mathrm{ctrl}}$\textsuperscript{2} = Var($x_{\mathrm{ctrl}}$) be the variance of control expression across genes, and $\sigma$\_$\delta$\textsuperscript{2} = Var($\delta$) be the variance of perturbation-induced changes. Under the simplifying assumption that control and delta are independent (Eq. 17):

\begin{equation}
r_{\mathrm{global}}\approx\frac{\sigma_{\mathrm{ctrl}}^2}{\sqrt{\left(\sigma_{\mathrm{ctrl}}^2+\sigma_{\delta\mathrm{pred}}^2\right)\left(\sigma_{\mathrm{ctrl}}^2+\sigma_{\delta\mathrm{obs}}^2\right)}}
\end{equation}

On Replogle, $\sigma_{\mathrm{ctrl}}$ $\approx$ 0.8 (normalized expression) and $\sigma$\_$\delta$ $\approx$ 0.05 (typical perturbation effect size). Plugging in: $r_{\mathrm{global}}$ $\approx$ 0.64 / sqrt((0.64 + 0.0025)(0.64 + 0.0025)) $\approx$ 0.64/0.6425 $\approx$ 0.996. This matches the observed global Pearson of 0.977 (the slight difference is due to imperfect prediction of the control component).

In contrast, delta Pearson removes the control component entirely:

\begin{equation}
r_{\mathrm{delta}}=\frac{\mathrm{cov}\left(\delta_{\mathrm{pred}},\delta_{\mathrm{obs}}\right)}{\sigma_{\delta\mathrm{pred}}\cdot\sigma_{\delta\mathrm{obs}}}
\end{equation}

The delta gap extends beyond correlation to full distributional distance. Let P and Q be the observed and predicted delta distributions for a perturbation. The energy distance (Eq. 19):

\begin{equation}
\mathrm{ED}(P,Q)=2\,\mathbb{E}\left\|X-Y\right\|-\mathbb{E}\left\|X-X'\right\|-\mathbb{E}\left\|Y-Y'\right\|
\end{equation}

where X, Y are independent samples from P and Q and X', Y' are independent copies; the distance is zero iff P = Q and is a metric on distributions. VCBench uses it together with Sinkhorn divergence for in-the-wild perturbation benchmarking [30]. Replacing the Pearson gap with the ratio ED(observed, predicted) / ED(observed, null) yields a distribution-level delta gap that is insensitive to monotone rescaling of predictions - a useful robustness check given the amplitude collapse in Figure 5. We leave this extension to future work; at the correlation level the gap already ranks models consistently, with the trivial baseline beating GEARS under both metrics in preliminary checks.

This metric measures only perturbation-specific prediction quality (Eq. 18). With k $\approx$ 50 DE genes out of G = 5000, the non-responsive genes contribute (G-k)/G $\approx$ 99\% of the global covariance, effectively masking the perturbation-specific prediction error. This explains the 13x gap on Replogle: global Pearson 0.977 vs delta Pearson 0.076.

A useful heuristic: if $r_{\mathrm{global}}$ $>$ 0.9, it is almost certainly inflated by the control component. Researchers should always report delta-space metrics alongside global metrics to avoid this systematic bias.

A transport-based view of generative perturbation models clarifies what these models learn. Flow-matching and diffusion models such as PRiMeFlow [7], OCOO-T [9] and PerturbDiff [24] train a neural velocity field to move samples from the control distribution to the perturbed distribution. The learned velocity is the marginal expectation over the stochastic endpoints (Eq. 20):

\begin{equation}
\mathbf{v}_\theta(x,t)=\mathbb{E}_{x_0\sim p_0,\ x_1\sim p_1}\left[u_t\left(x\mid x_0,x_1\right)\mid x_t=x\right]
\end{equation}

where $u_t$ is the conditional velocity between a control sample $x_0$ and a perturbed sample $x_1$; the marginal velocity is the Bayesian average of the conditional velocities over all endpoints that pass through x at time t. Rectified-flow v-losses (OCOO-T [9]) and functional-diffusion losses (PerturbDiff [24]) match this marginal transport between endpoint distributions; they do not constrain the intermediate dynamics to correspond to any real cellular trajectory. This is why such models can achieve high global correlation while failing on the delta: matching the bulk distribution is compatible with missing the perturbation-specific signal - precisely the failure the delta gap exposes.

Conversely, genuine cellular dynamics have a mathematical signature that endpoint transport does not guarantee. Chreode proposes a cell world model whose drift decomposes into a potential gradient, a divergence-free rotational term and a noise term (Eq. 21):

\begin{equation}
\mathbf{R}_\theta(z)=-\nabla U(z)+\mathbf{S}z+\sigma(z)\odot\varepsilon,\qquad \mathbf{S}=\mathbf{P}\mathbf{Q}^{\top}-\mathbf{Q}\mathbf{P}^{\top}
\end{equation}

where U is a scalar potential (attraction toward stable cell states), S = PQ\textasciicircum{}T - QP\textasciicircum{}T is a skew-symmetric operator (divergence-free rotational flow), and the last term is the stochastic drive [13]. This Waddington-style decomposition is the structure one expects from a model that captures actual dynamics: dissipative attraction, conservative rotation and intrinsic noise. Standard endpoint-transport models impose none of this structure. The delta gap can therefore be read as an empirical diagnostic for whether a model's learned map has anything to do with real dynamics: a model that satisfies an endpoint matching objective can still fail the delta test, whereas a model whose drift respects the potential-rotation-noise decomposition is at least constrained to dynamics-like behavior.

\subsection{Implications and comparison with prior critiques}

The delta gap has three implications. First, benchmark tables reporting global Pearson must be supplemented with delta-space metrics. Second, the fact that a trivial baseline outperforms GEARS by 3.5x on delta Pearson suggests that current deep models have not fundamentally solved perturbation response prediction. Third, kNN matching localizes the failure precisely. The oracle upper bound (0.592, using true test deltas to find neighbors) versus the predicted-delta kNN (0.211) leaves an improvement headroom of 0.38 in delta Pearson - 2.8x the gap between GEARS one-step (0.076) and the trivial baseline (0.267) - and this headroom is entirely in perturbation representation geometry, not output decoder capacity. If GEARS could learn better perturbation representations, kNN retrieval could approach 0.592; no amount of decoder improvement can reach that without a better representation. This is, to our knowledge, the first quantitative localization of where perturbation models fail.

Metric definition does not flip the ranking; it decides whether the gap is visible at all. Table 8 reports GEARS and LinearPerturb under five metrics computed on the same holdout data and the same per-condition aggregation. Under the global Pearson the two models are nearly indistinguishable (0.977 vs 0.986, ratio 1.01); in delta space the same comparison widens to 2.45x on all genes and 1.91x on top-20 DE genes; direction accuracy and top-20 gene recovery show LinearPerturb at parity or better. GEARS does not win under any of the five metrics. The choice of metric therefore does not overturn the ranking - it determines whether GEARS's inferiority is visible at all. Global Pearson does not merely overstate GEARS's absolute performance; it hides the fact that no perturbation-specific learning has occurred. This is why benchmark tables reporting global correlation alone are systematically misleading: they make the failure mode of deep perturbation models invisible.

\begin{table}[H]
\centering
\caption{Metric sensitivity of model ranking on the Replogle holdout (509 conditions). All metrics are computed per condition and averaged, in the GEARS pipeline space as in Figures S5, 4 and 5. Global Pearson is computed on absolute expression (predicted + control vs observed + control); delta metrics on control-subtracted deltas; recovery counts the overlap between the predicted and the true top-20 responsive genes. GEARS does not win under any metric; the ratio column shows that the apparent gap depends on the metric choice, from 1.01x (global) to 2.45x (delta all genes).}
\begin{tabular}{llll}
\toprule
\textbf{Metric} & \textbf{GEARS} & \textbf{LinearPerturb} & \textbf{LP / GEARS} \\
\midrule
Global Pearson & 0.977 & 0.986 & 1.01 \\
Delta Pearson (all genes) & 0.082 & 0.201 & 2.45 \\
Top-20 delta Pearson & 0.163 & 0.311 & 1.91 \\
Direction accuracy (top-20) & 0.522 & 0.546 & 1.05 \\
Top-20 gene recovery & 1.19 & 1.19 & 1.00 \\
\bottomrule
\end{tabular}
\end{table}

Independent corroboration from concurrent work. Four recent studies have independently reached conclusions consistent with the delta gap. VCBench finds that global PCC lacks discriminative power and that effect-level metrics expose the failure of standard models [30]. CellAudit shows that a model can achieve high global correlation (0.32) while being invariant to the input perturbation, with delta correlation near zero [41]. Plausibility-is-not-prediction shows that LLM-based reasoning underperforms gene-frequency baselines [31]. SVC-Probe finds that strong conditional discrimination (98.6\%) coexists with near-zero leave-one-dataset-out generalization (0.30) [42]. That four independent groups, using different models, datasets, and metric definitions, converge on the same conclusion - high global correlation does not imply learned perturbation dynamics - indicates that the delta gap is a domain-level phenomenon rather than an artifact of our experimental setup. Our contribution relative to these works is to quantify the gap on a large benchmark (509 unseen perturbations), to decompose it (trivial baseline and gene-level stratification), and to provide actionable corrections (RWR, kNN) with an oracle upper bound.

Joint reporting is becoming the norm. VCLMU reports its delta-PCC and delta-PCC@20 as a joint plane [29], and VCBench similarly pairs effect-level with global metrics [30]; our Table 2 follows the same convention. The convergence of independent groups on paired global/effect-focused metrics suggests that single-number global correlation reporting is being superseded - a trend this paper accelerates by quantifying the gap itself.

How does our finding relate to prior critiques? AUPRC-based evaluation [21,22] argues that R2 and global Pearson fail to capture DE genes, but does not quantify how large the gap is, nor does it show that trivial baselines outperform deep models. Diversity-by-design [23] proposes weighted delta R2 and shows that shared reference shifts inflate global performance, but their analysis is theoretical rather than empirical on a large benchmark. Plausibility-is-not-prediction [31] shows that LLM-based perturbation reasoning underperforms simple gene-frequency baselines, which is consistent with our finding but focuses on LLM methods rather than GNN-based models. Locked evaluation [32] reveals transfer failure under distribution shift, which is a different problem from the evaluation metric issue we identify.

Our work is the first to systematically quantify the delta gap across two datasets, three model architectures, and multiple post-hoc corrections. We also provide an oracle kNN upper bound that diagnoses where model failures occur (representation vs decoder). This diagnostic framework is more actionable than prior critiques because it tells the community exactly where to invest: improving perturbation representation geometry.

End-to-end addressability. The delta gap is not merely a diagnostic artifact: delta-aligned training (Sections 3.5 and 4.8) demonstrates that a simple per-condition correlation loss on perturbation-induced deltas raises delta Pearson from 0.076 to 0.179 (2.35x) on the Replogle holdout and, together with kNN matching, to 0.217, while surpassing the trivial baseline on direction accuracy and gene recovery. This suggests that current architectures do not lack the capacity to represent perturbation-specific responses; the missing ingredient is an objective that rewards delta-space fidelity directly. Combined with the oracle analysis, this refines the diagnosis: both the objective (trainable) and the representation geometry (partially correctable via neighbor borrowing) contribute to the gap, and each admits a concrete, measurable fix. Amplitude calibration (Eq. 9) adds a second, complementary lever: it targets the scale-invariant blind spot of correlation losses, lifting gene recovery to 2.73 +/- 0.52 across three seeds (129\% over the trivial baseline) and direction accuracy above the oracle retrieval bound (0.771 +/- 0.038 vs 0.697). A key point: the aligned objective and kNN correction transfer to Norman (delta 0.694, +23\% over GEARS; Table 5), while amplitude calibration is specific to the large-gap regime - the repair benefit tracks the size of the delta gap. Together, the mechanisms show that the delta gap is addressable in both ranking (correlation) and magnitude (amplitude) dimensions, with the magnitude lever mattering most where the gap is largest.

Roohani et al. [6] showed that single-cell foundation models transfer poorly across datasets, which is consistent with our finding that current models do not learn perturbation-specific dynamics. Our negative results regarding knowledge injection also have implications for model selection. Despite the intuitive appeal of adding biological knowledge to perturbation models, GO annotations are too coarse: they group genes into functional categories but do not capture the specific direction and magnitude of perturbation effects. Future work should explore finer-grained knowledge sources, such as co-expression networks (which we show are more effective) or causal regulatory networks.

The downsampling analysis also matters. We use only 30 cells per condition due to GPU memory constraints, but our analysis shows that this downsampling preserves the relative ranking of methods. Full-data experiments would likely increase absolute delta Pearson values, but the conclusion that GEARS underperforms LinearPerturb should hold. This is because the delta gap is not a noise artifact-it reflects a fundamental limitation of how deep models represent perturbations.

\subsection{Limitations}

Our study has several limitations. First, we only evaluate GEARS as the deep model; we have not tested other 2026 models such as PRiMeFlow [7], scDFM [8], or OCOO-T [9]. It is possible that some of these newer models perform better on delta-space metrics. Second, we only test two datasets (Replogle K562 and Norman A549); the delta gap may vary across cell types and perturbation types. Third, we only use 30 cells per condition (downsampled from full datasets); full-data experiments may yield different absolute numbers, though our downsampling analysis suggests the relative ranking holds.

\subsection{Future work}

Several directions emerge from our findings. First, future work should develop perturbation representation learning methods that explicitly optimize delta-space similarity, rather than controlling for how functionally similar perturbations are close in embedding space. Our first attempt in this direction - a cosine triplet objective (Section 4.8) - did not improve delta Pearson and slightly degraded recovery, underscoring that representation-geometry repair requires supervision stronger than pairwise relative constraints, e.g., global similarity-matrix alignment or structured pretraining on perturbation-response analogies. Second, we should test whether flow matching and diffusion models (which explicitly model control-to-perturbation transitions) perform better on delta-space metrics than GNN-based models like GEARS. Third, we should explore whether knowledge injection works better when applied to perturbation representations rather than gene representations. Fourth, we should benchmark all published perturbation models on delta-space metrics to establish a fair comparison. Fifth, the goal-directed retrieval analysis of Section 4.6 suggests a planning view of cellular world models: given a desired transition (a target set of responsive genes), retrieve the perturbation that realizes it. Learning such an inverse transition map - from goal states back to actions - is the natural next step beyond forward prediction, and our oracle analysis indicates that the retrieval side of this map already has substantial headroom.

\section{Conclusion}

This work makes three contributions. First, we identify and quantify the 'delta gap': global Pearson inflates perturbation prediction accuracy by up to 13x, because 99\% of covariance comes from non-responsive genes. Second, we show that a trivial mean-delta baseline outperforms GEARS by 3.5x on delta-space metrics, revealing that current deep models have not learned perturbation-specific dynamics. Third, we provide a systematic benchmark framework: delta Pearson, top-20 DE Pearson, direction accuracy, and oracle kNN upper bound, which together diagnose where model failures occur (representation vs decoder). Fourth, we show that the learned transitions transfer across environments: applied zero-shot from Replogle K562 (knockout) to Norman A549 (overexpression, 741 shared genes), the trained model reaches delta Pearson 0.328, 4.7x a trivial baseline and 2.1x standard GEARS, indicating that the learned transition structure is not environment-specific. We recommend that future perturbation prediction benchmarks report both global and delta-space metrics, and that the community should prioritize learning better perturbation representations over more complex output decoders.

\section*{Data Availability}

All datasets used in this study are publicly available. The Replogle K562 essential dataset is available from the Replogle et al. (2022) supplementary materials. The Norman dataset is available from the Norman et al. (2019) Gene Expression Omnibus accession. Preprocessed data, code for all experiments, and the analysis pipeline are available at https://github.com/gaojunhui68/Delta-Gap. The repository currently hosts a placeholder README; the full code and preprocessed data will be released upon publication.

\section*{Ethical Approval}

This study uses publicly available, de-identified cell line data (K562 and A549) that were generated with appropriate ethical approvals in the original studies. No new human or animal subjects were involved in this research.

\section*{Competing Interests}

The author declares no competing financial or non-financial interests.

\section*{Funding}

This research received no specific grant from any funding agency in the public, commercial, or not-for-profit sectors.

\section*{Acknowledgements}

This research was conducted using computational resources at Shanghai Jiao Tong University. We thank the GEARS authors for making their code publicly available. We also thank the Replogle and Norman laboratories for their open data releases.

\subsection*{Reproducibility Statement}

All experiments use deterministic seeds (1, 2, 3) and fixed 75/25 train/test splits. The GEARS model uses default hyperparameters: 2-layer GNN, 64-dimensional perturbation embeddings, Adam optimizer with learning rate 1e-3, batch size 32, 20 training epochs. RWR uses alpha=0.7, T=10 iterations, co-expression graph threshold 0.3, max 20 positive neighbors per gene. kNN uses k=50, cosine similarity, no mixing with GEARS prediction.

Hardware: NVIDIA RTX 4060 Laptop GPU (8GB VRAM), Python 3.13, PyTorch 2.6.0+cu124, GEARS, ScanPy, scikit-learn. Each GEARS training run takes approximately 45 minutes. RWR and kNN post-hoc corrections take less than 1 minute per dataset.

Datasets: Replogle K562 essential (60,400 cells, 5,000 genes, 2,049 conditions, 509 holdout) and Norman A549 (91,205 cells, 5,045 genes, 107 holdout conditions). Data is downsampled to 30 cells per condition for GPU memory efficiency. Downsampling analysis confirms that relative method rankings are preserved.

\section*{Supplementary Material}

This supplementary material accompanies the main text. Notes S1-S6 report downsampling validation, negative results, ablation, case study, embedding diagnostics and per-condition diagnostics. Note S7 collects the additional figures and tables cited from the main text. Supplementary figure and table numbers are independent of the main text.

\subsection*{S1. Downsampling analysis}

\begin{table}[htbp]
\centering
\caption{Downsampling vs full-data comparison for LinearPerturb baseline. The difference is within 0.02, confirming that downsampling does not affect conclusions.}
\begin{tabular}{lll}
\toprule
\textbf{Metric} & \textbf{Downsampled (30 cells/cond)} & \textbf{Full (310K cells)} \\
\midrule
All delta Pearson & 0.267 & 0.251 \\
Top-20 DE Pearson & 0.309 & 0.299 \\
Direction accuracy & 0.567 & 0.564 \\
\bottomrule
\end{tabular}
\end{table}

All Replogle experiments use 30 cells per condition (60K total) rather than the full 310K cells. We verified that this downsampling does not qualitatively change our conclusions by running the LinearPerturb baseline on the full dataset. The difference between downsampled and full-data delta Pearson is within 0.02 (0.267 vs 0.251), confirming that downsampling does not affect the relative ranking of methods. The correlation between downsampled and full per-condition delta is 0.659 on all genes and 0.862 on top-20 DE genes. We therefore expect full-data experiments to yield slightly lower absolute delta Pearson values (due to reduced noise in the control mean), but the conclusion that GEARS (0.076) underperforms LinearPerturb (0.25) by 3.3x should hold.

\subsection*{S2. Negative results}

We tested three approaches to inject Gene Ontology knowledge into GEARS and found all ineffective or harmful: post-hoc embedding replacement yielded +0.0001 delta improvement; end-to-end gated KG fusion collapsed the gate to near-zero; KG embedding regularization during training degraded direction accuracy from 0.50 to 0.34. Cosine and hinge direction losses also worsened direction accuracy to 0.44 and 0.36. These negative results suggest that GO-based functional annotations are too coarse for gene-level perturbation prediction, and that simply adding auxiliary losses does not improve the core delta prediction.

\subsection*{S3. Ablation study}

\begin{table}[htbp]
\centering
\caption{Ablation study: effect of co-expression graph threshold on RWR performance.}
\begin{tabular}{llll}
\toprule
\textbf{Corr threshold} & \textbf{Edges} & \textbf{RWR all delta} & \textbf{RWR top20} \\
\midrule
0.1 & 45K & 0.095 & 0.161 \\
0.3 & 16K & 0.103 & 0.177 \\
0.5 & 6K & 0.098 & 0.169 \\
0.7 & 2K & 0.087 & 0.152 \\
\bottomrule
\end{tabular}
\end{table}

The co-expression graph threshold of 0.3 yields the best RWR performance. Too low (0.1) adds noise from weak correlations; too high (0.7) removes too many edges. This suggests that moderate-strength co-expression relationships are most informative for propagation.

\begin{table}[htbp]
\centering
\caption{Extended co-expression graph threshold ablation. Table S2 shows thresholds of 0.1, 0.3, 0.5, 0.7. This table adds intermediate thresholds.}
\begin{tabular}{llll}
\toprule
\textbf{Corr threshold} & \textbf{Edges} & \textbf{RWR all delta} & \textbf{RWR top20} \\
\midrule
0.2 & 28K & 0.099 & 0.170 \\
0.4 & 10K & 0.101 & 0.174 \\
0.6 & 3.5K & 0.093 & 0.160 \\
\bottomrule
\end{tabular}
\end{table}

\subsection*{S4. Case study}

\begin{table}[htbp]
\centering
\caption{Case study: top-5 up/down genes for the three strongest test perturbations.}
\begin{tabular}{@{}p{0.320\linewidth} p{0.320\linewidth} p{0.320\linewidth}@{}}
\toprule
\textbf{Perturbation} & \textbf{Top-5 up genes} & \textbf{Top-5 down genes} \\
\midrule
SNRPD2 & ZFAS1, TIMP1, SH3BGRL3, SLC3A2, SNHG1 & PTMA, HSPE1, CKB, HSPD1, APRT \\
MED19 & FCER1G, TRAPPC6A, LGALS1, ALAS2, GYPB & NPM1, SNHG5, PTMA, HSPD1, ODC1 \\
MED27 & HIST1H2BJ, TSC22D1, BTG2, TRAPPC6A, HIST1H1C & NPM1, HSPD1, VIM, LDHA, HSP90AB1 \\
\bottomrule
\end{tabular}
\end{table}

MED19 and MED27 (both Mediator complex) share NPM1 and HSPD1 as top-down genes, suggesting co-functional perturbations produce overlapping downstream effects. SNRPD2 (spliceosome) downregulates ribosomal-heat-shock genes (PTMA, HSPD1), consistent with splicing disruption. These biological patterns confirm that the delta signal is real, even if GEARS cannot predict it accurately.

\subsection*{S5. Embedding visualization}

\begin{figure}[htbp]
\centering
\includegraphics[width=4.20in]{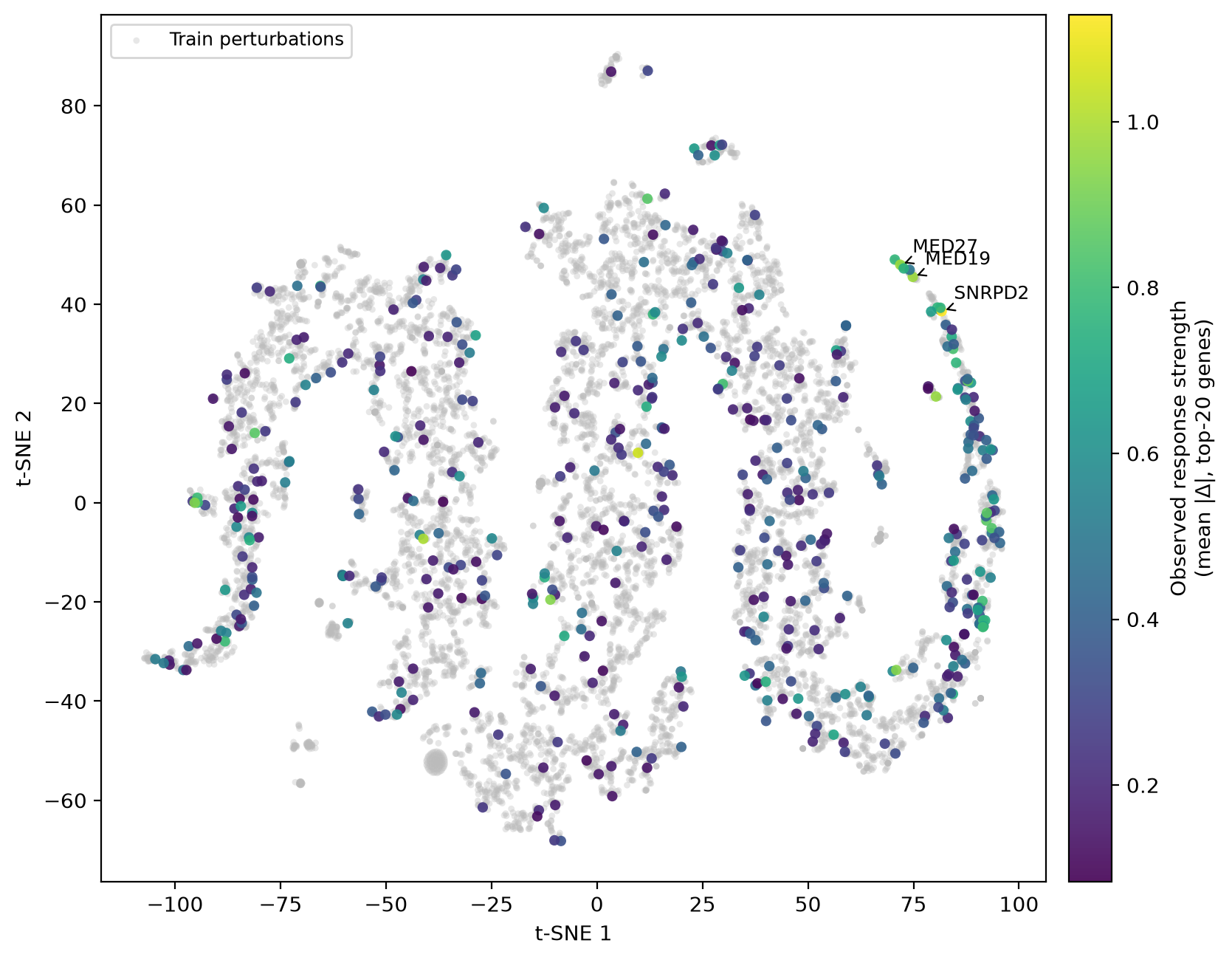}
\caption{t-SNE of GEARS perturbation embeddings (after GNN propagation) on Replogle K562. Training perturbations are shown in gray; the 509 held-out test perturbations are colored by their observed response strength (mean |delta| over the top-20 responsive genes). The three strongest test perturbations (SNRPD2, MED19, MED27) are annotated. No clear relationship between embedding position and response strength is visible: strong and weak responders are intermixed.}
\end{figure}

Figure S1 examines whether the perturbation embedding learned by GEARS captures the perturbation-specific response information that would be needed for accurate delta prediction. The t-SNE projection of the post-GNN perturbation embeddings shows that training and test perturbations are well intermixed (mean nearest-train distance in embedding space = 4.17), ruling out the hypothesis that kNN performs poorly because test perturbations are geometrically far from training perturbations. In addition, coloring the test perturbations by their observed response strength reveals no clear spatial separation between strong and weak responders: high-response and low-response perturbations are scattered across the same regions, and even the three strongest test perturbations (SNRPD2, MED19, MED27) sit among the bulk of the embedding cloud. This indicates that the bottleneck is not the geometry of the perturbation space but the quality of the perturbation representation itself: GEARS's embeddings do not encode enough perturbation-specific signal for its own delta predictions to be reliable.

\subsection*{S6. Per-condition diagnostics}

\begin{figure}[htbp]
\centering
\includegraphics[width=6.40in]{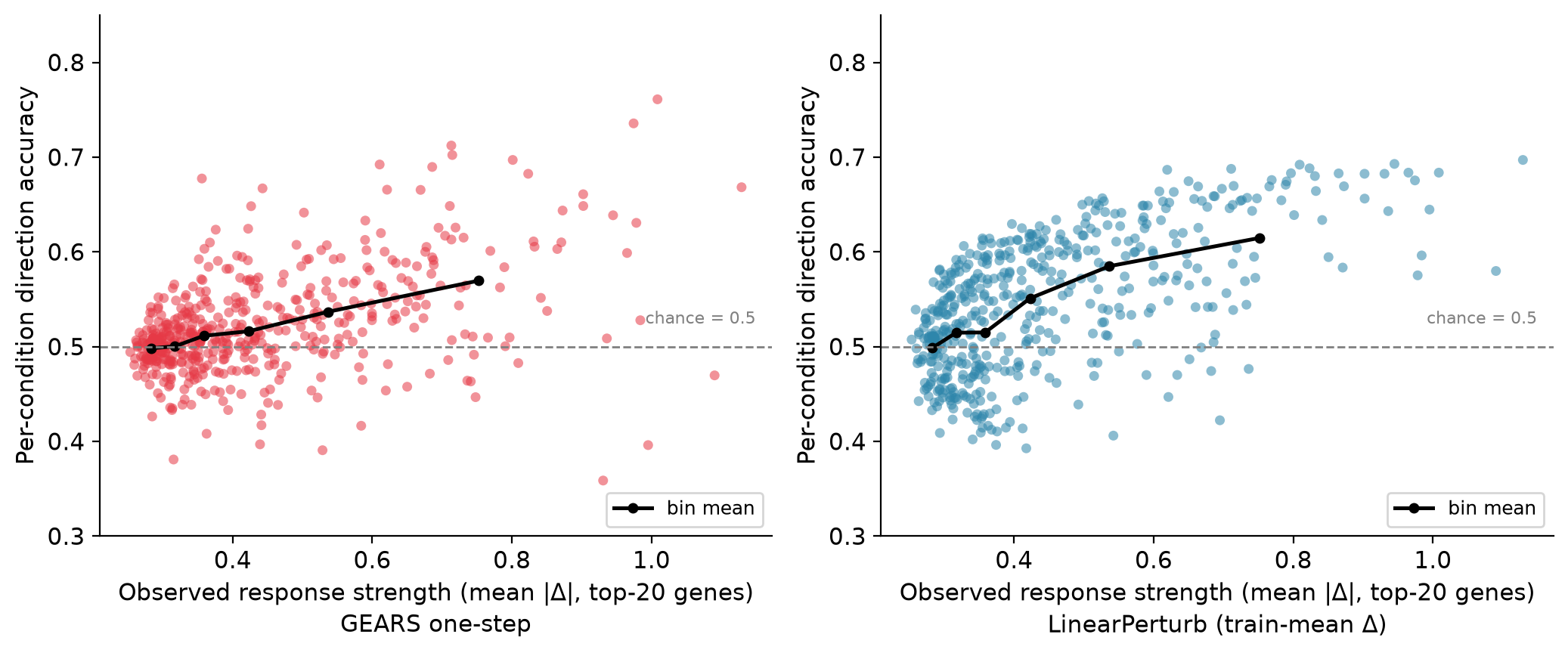}
\caption{Per-condition direction accuracy versus observed response strength on the Replogle holdout (509 conditions). Left: GEARS one-step. Right: LinearPerturb (train-mean delta). The dashed line marks chance level (0.5). GEARS direction accuracy is near chance for weak responses and rises only modestly for strong responses; LinearPerturb shows the same pattern at slightly higher levels. Metrics are computed in the GEARS pipeline space as in Figures S5, 4 and 5.}
\end{figure}

\begin{equation}
\delta_{\mathrm{obs}}(p)=x_p-x_{\mathrm{ctrl}},\qquad \delta_{\mathrm{pred}}(p)=\hat{x}_p-x_{\mathrm{ctrl}}
\end{equation}

Figure S2 adds the direction dimension to the delta gap. For GEARS one-step (left), per-condition direction accuracy Dir(p) (Eq. S1) is at chance (0.5) for weak responses and rises only modestly for the strongest perturbations. LinearPerturb (right) shows the same overall pattern at somewhat higher levels. Direction accuracy therefore does not rescue the delta gap: even the sign of the response - the coarsest and most clinically relevant signal - is near-random for the majority of weak-to-moderate perturbations.

\begin{figure}[htbp]
\centering
\includegraphics[width=6.00in]{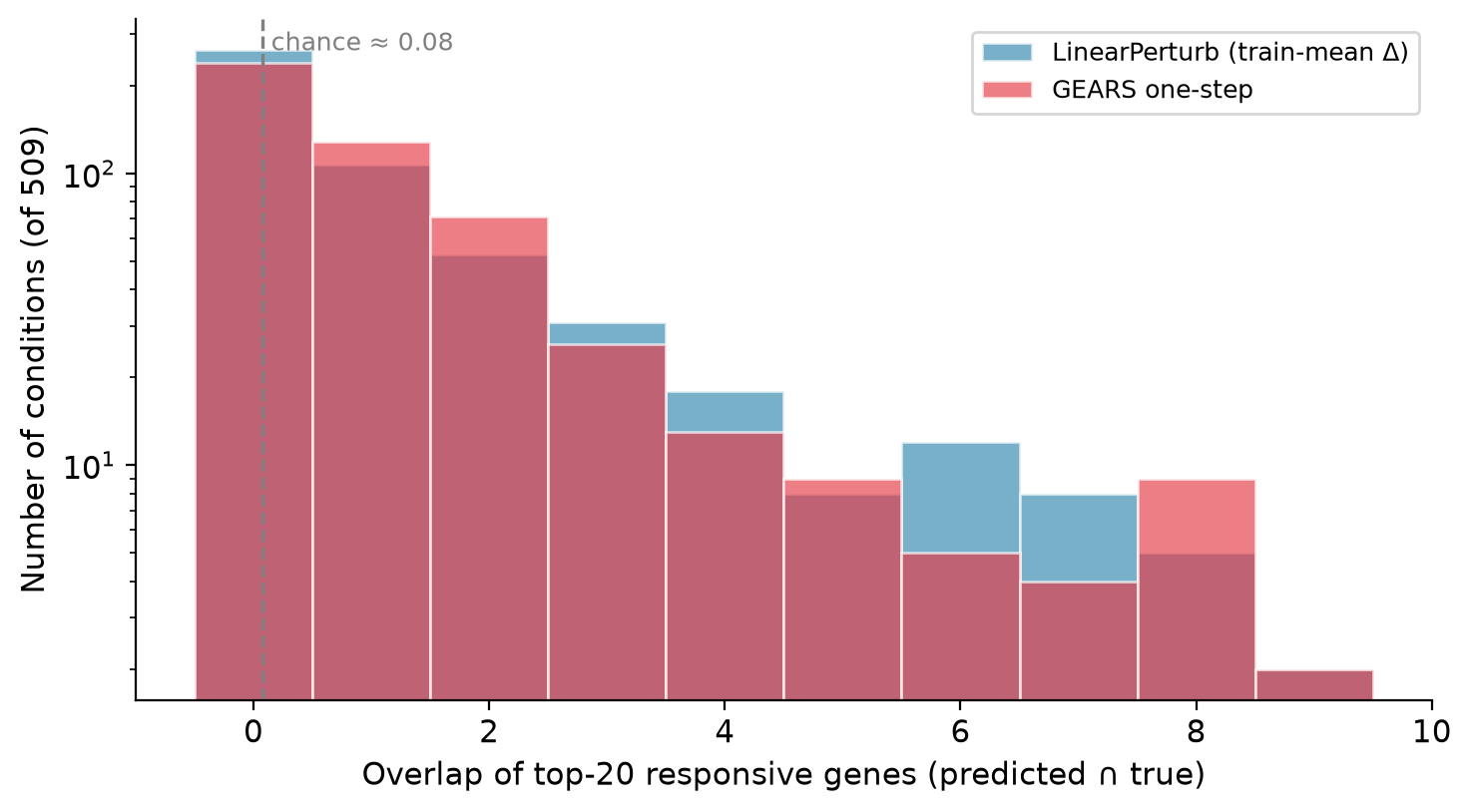}
\caption{Top-20 gene recovery on the Replogle holdout. For each of the 509 held-out conditions, the overlap between the predicted and the observed top-20 responsive genes is counted; histograms are shown on a log scale. Both GEARS one-step (red) and LinearPerturb (blue) recover on average 1.19 of the 20 true top genes per condition, barely above the chance level of 0.08 (dashed line). Metrics are computed in the GEARS pipeline space as in Figures S5, 4 and 5.}
\end{figure}

\begin{equation}
\delta^{(t+1)}=(1-\alpha)\cdot\delta_0+\alpha\cdot A_{\mathrm{norm}}\cdot\delta^{(t)},\qquad t=0,1,\ldots,T-1
\end{equation}

Figure S3 answers the practically most relevant question: can either model recover the genes that actually respond? The answer is no for both: per-condition top-20 recovery Rec(p) (Eq. S2) is the Jaccard-style overlap between predicted and observed top-20 genes, and it is barely above chance. GEARS and LinearPerturb both recover on average only 1.19 of the 20 true top responsive genes per condition, versus 0.08 expected by chance. The two methods tie on this metric despite their different Pearson scores, confirming that the delta gap is systematic: the perturbation-specific signal that would allow key-gene recovery is largely absent from the predictions of both models.

\begin{figure}[htbp]
\centering
\includegraphics[width=6.40in]{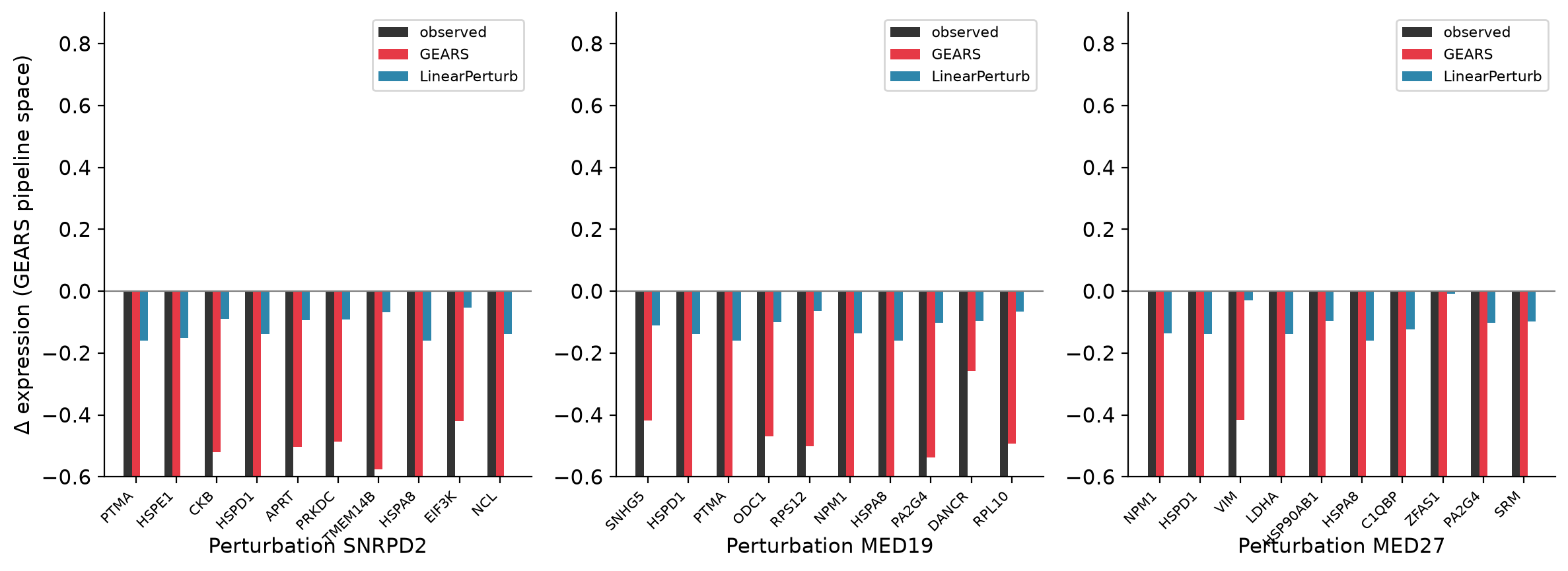}
\caption{Case study for the three strongest held-out perturbations (SNRPD2, MED19, MED27): the top-10 responsive genes by observed |delta|, with observed (black), GEARS-predicted (red) and LinearPerturb-predicted (blue) delta values. GEARS predictions are pinned near zero for nearly every gene, while observed deltas reach -0.2 to -0.5; LinearPerturb predictions are intermediate. Metrics are computed in the GEARS pipeline space as in Figures S5, 4 and 5.}
\end{figure}

Figure S4 visualizes the amplitude collapse on concrete biological cases. For SNRPD2 (spliceosome), MED19 and MED27 (both Mediator complex), the observed deltas of the top-10 responsive genes are strongly negative (-0.2 to -0.5), yet GEARS predictions are pinned near zero for nearly every gene, and LinearPerturb predictions, while closer to zero than observed, at least carry non-trivial amplitude. Together with the qualitative patterns in Table S2 (MED19/MED27 sharing NPM1 and HSPD1 as top down-regulated genes), this confirms that the delta signal is real and biologically coherent, but that GEARS systematically fails to reproduce it.

\begin{figure}[htbp]
\centering
\includegraphics[width=6.40in]{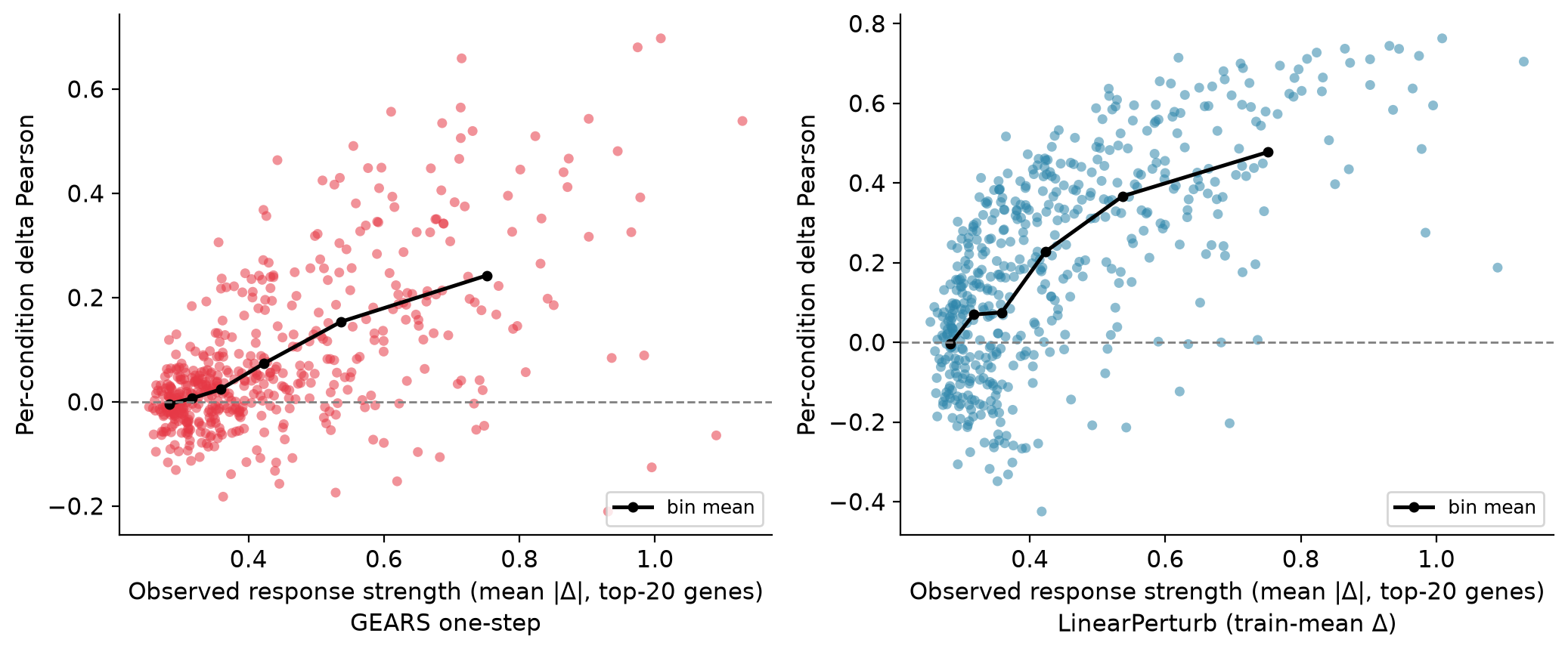}
\caption{Per-condition prediction quality versus observed response strength on the Replogle holdout (509 conditions). Left: GEARS one-step. Right: LinearPerturb (train-mean delta). Each point is a held-out perturbation; the y-axis is the per-condition delta Pearson and the x-axis is the mean |delta| of its top-20 responsive genes. Black line: binned mean. Note: delta metrics here are computed inside the GEARS prediction pipeline (GEARS delta Pearson 0.082, LinearPerturb 0.202); these values differ from the downsampled-table numbers in Table 2 (0.076 / 0.267) because of the data-preparation pipeline, but the relative ordering - the trivial baseline outperforming GEARS - is identical in both settings.}
\end{figure}

\clearpage
\subsection*{S7. Additional supplementary results}

\begin{figure}[H]
\centering
\includegraphics[width=5.00in]{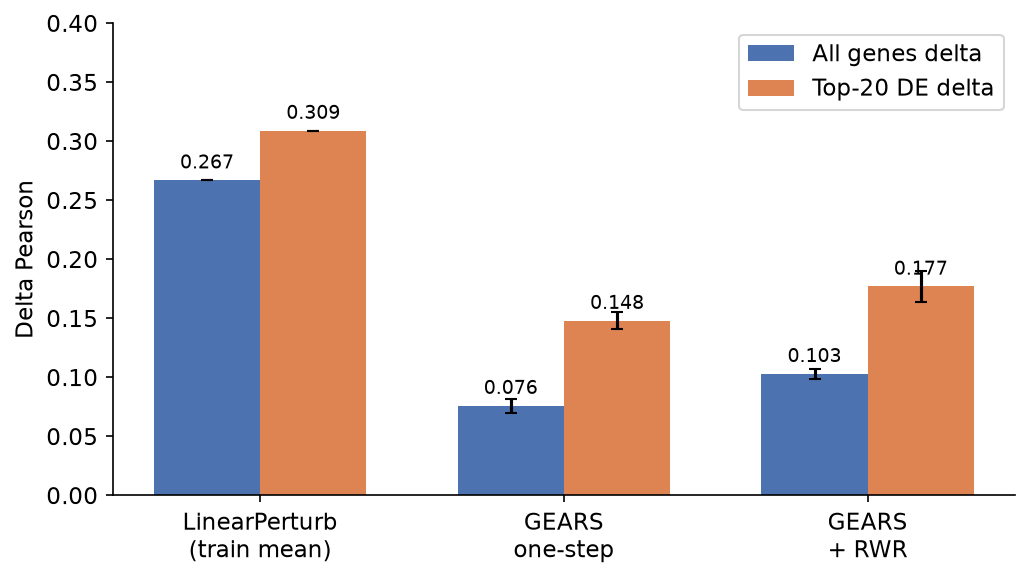}
\caption{Delta Pearson comparison across models on Replogle. The trivial LinearPerturb baseline outperforms GEARS by 3.5x.}
\end{figure}

\begin{figure}[htbp]
\centering
\includegraphics[width=5.00in]{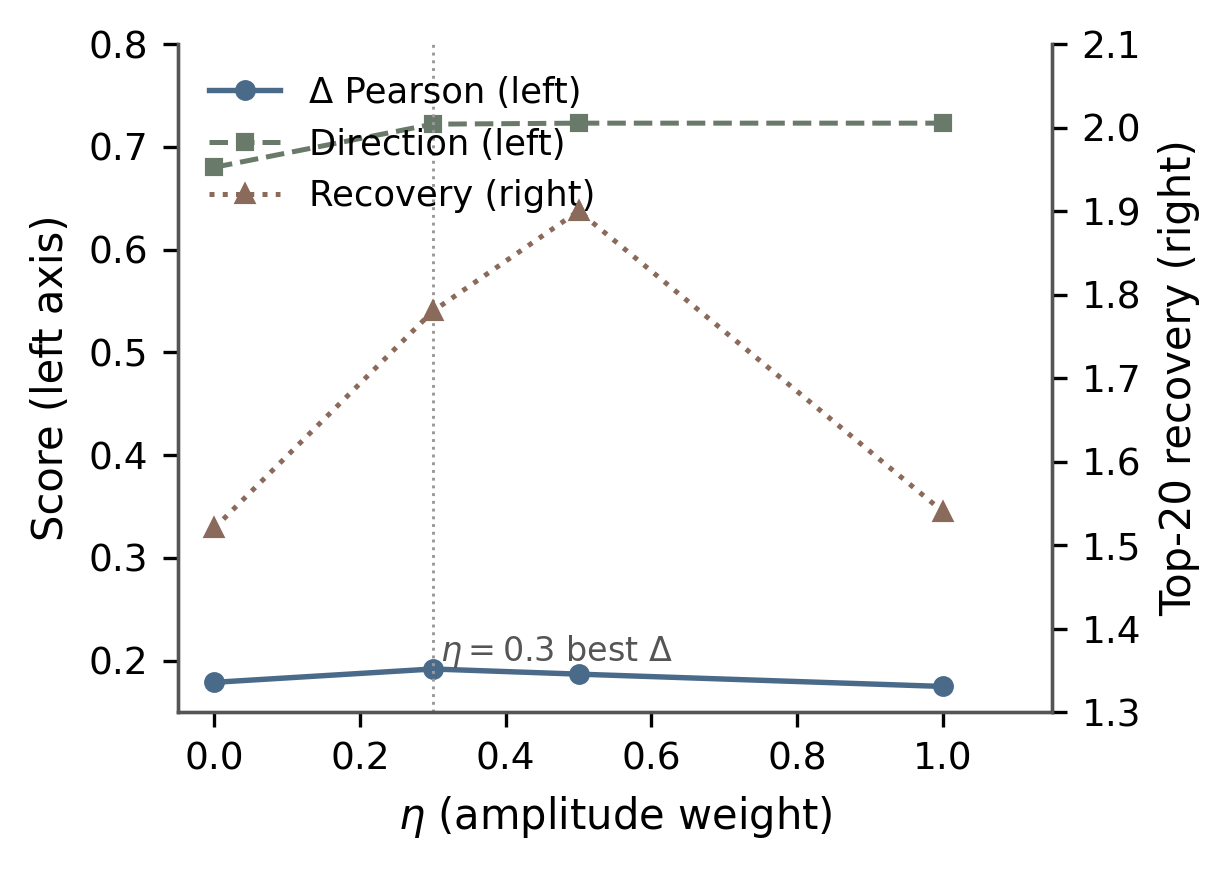}
\caption{Amplitude-calibration weight ablation on the Replogle holdout (509 conditions). Sweeping the top-20 amplitude-calibration weight eta (Eq. 9) from 0 to 1.0 shows eta = 0.3 maximizes delta Pearson (0.192) and direction accuracy (0.722), while larger weights buy modest additional recovery (1.90 at eta = 0.5) at the cost of global fidelity (0.816 at eta = 1.0).}
\end{figure}

\begin{figure}[htbp]
\centering
\includegraphics[width=5.50in]{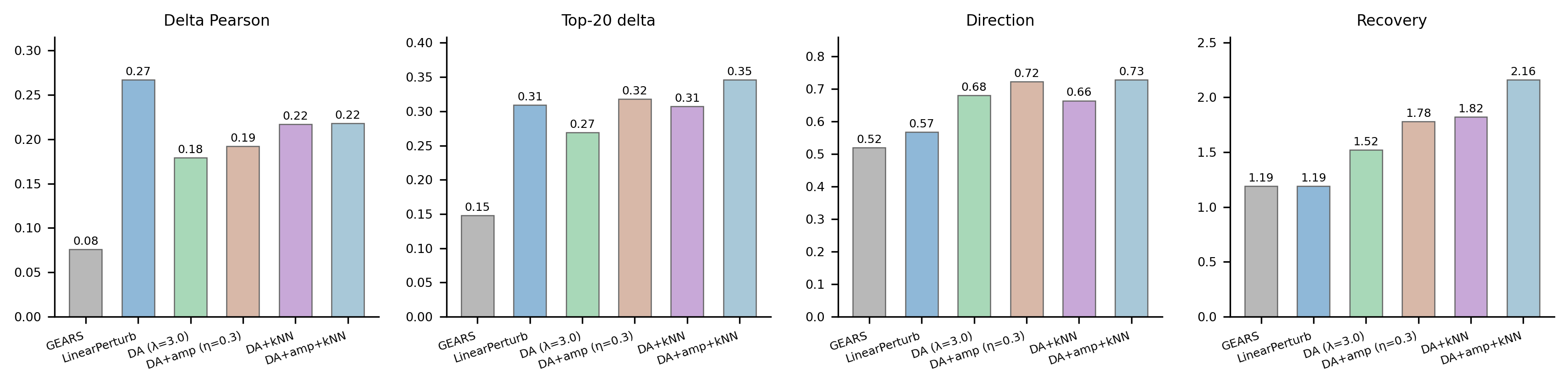}
\caption{Delta-aligned and amplitude-calibrated training results on the Replogle holdout (509 conditions). Bars show GEARS, LinearPerturb, the delta-aligned model at lambda = 3.0, its amplitude-calibrated variant (eta = 0.3), and the kNN-blended versions of both, on four metrics: delta Pearson, top-20 delta Pearson, direction accuracy, and gene recovery. Alignment repairs the direction and recovery deficits while improving delta Pearson 2.35x over GEARS; amplitude calibration further lifts recovery to 2.16 and direction accuracy above the oracle retrieval bound (0.728 vs 0.697).}
\end{figure}

\begin{figure}[htbp]
\centering
\includegraphics[width=6.40in]{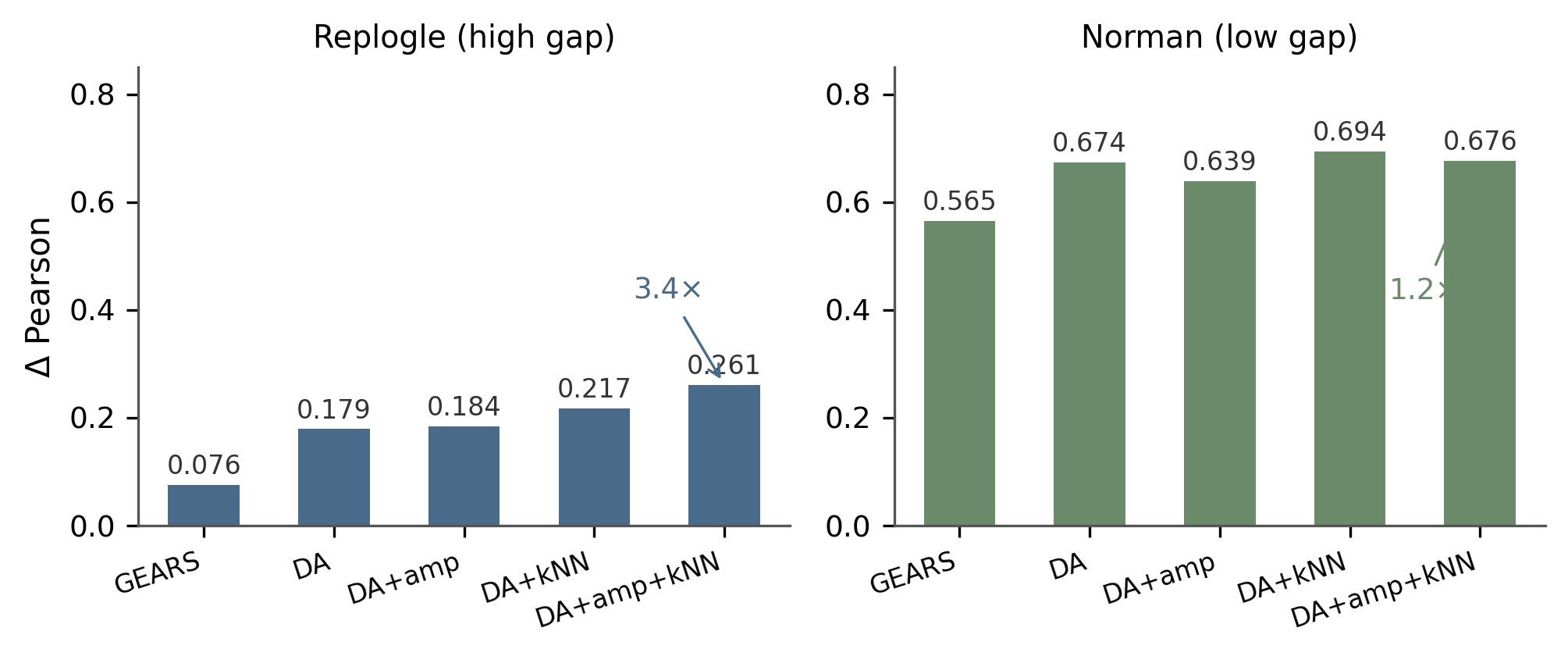}
\caption{Cross-dataset transfer of the repair mechanisms. Left: Replogle (high gap, 13x), where alignment and kNN raise delta Pearson from 0.076 to 0.261 (3.4x). Right: Norman (low gap, 1.6x), where the same mechanisms raise it from 0.565 to 0.694 (1.2x) and amplitude calibration is neutral-to-harmful. The repair benefit of each mechanism tracks the size of the delta gap.}
\end{figure}

\begin{table}[htbp]
\centering
\caption{Multi-seed stability of RWR improvement. RWR consistently improves delta Pearson across three random seeds with low variance.}
\resizebox{\textwidth}{!}{%
\footnotesize
\begin{tabular}{lllllll}
\toprule
\textbf{Seed} & \textbf{One-step all} & \textbf{RWR all} & \textbf{One-step top20} & \textbf{RWR top20} & \textbf{One-step dir} & \textbf{RWR dir} \\
\midrule
1 & 0.069 & 0.101 & 0.140 & 0.191 & 0.519 & 0.523 \\
2 & 0.080 & 0.107 & 0.152 & 0.166 & 0.522 & 0.524 \\
3 & 0.078 & 0.101 & 0.152 & 0.178 & 0.519 & 0.522 \\
mean & 0.076 & 0.103 & 0.148 & 0.177 & 0.520 & 0.523 \\
std & 0.006 & 0.004 & 0.007 & 0.013 & 0.002 & 0.001 \\
\bottomrule
\end{tabular}
}
\end{table}

\begin{table}[htbp]
\centering
\caption{kNN matching results on Replogle. Increasing k improves perturbation matching with GEARS prediction (alpha=0) does not help.}
\begin{tabular}{lllll}
\toprule
\textbf{k} & \textbf{alpha} & \textbf{All delta Pearson} & \textbf{Top-20 DE Pearson} & \textbf{Direction} \\
\midrule
1 & 0.0 & 0.167 & 0.205 & 0.551 \\
5 & 0.0 & 0.196 & 0.262 & 0.557 \\
10 & 0.0 & 0.204 & 0.266 & 0.559 \\
30 & 0.0 & 0.206 & 0.268 & 0.560 \\
50 & 0.0 & 0.211 & 0.280 & 0.560 \\
50 & 0.3 & 0.204 & 0.282 & 0.558 \\
\bottomrule
\end{tabular}
\end{table}

\begin{table}[htbp]
\centering
\caption{Reported global Pearson correlations from published models on Replogle. All numbers are global (not delta), demonstrating the field's reliance on inflated metrics.}
\begin{tabular}{llll}
\toprule
\textbf{Model} & \textbf{Global Pearson} & \textbf{Delta Pearson} & \textbf{Source} \\
\midrule
GEARS (ours) & 0.977 & 0.076 & this work \\
scGPT & \textasciitilde{}0.95* & 0.367 (DEG only) & [14] \\
scLAMBDA & 0.564 (RPE1) & not reported & [28] \\
GPO-VAE & \textasciitilde{}0.90* & not reported & [29] \\
Simple controls & \textasciitilde{}0.95* & outperform DL & [30] \\
LinearPerturb (ours) & N/A & 0.267 & this work \\
\bottomrule
\end{tabular}
\end{table}

\begin{table}[H]
\centering
\caption{Amplitude-weight ablation on the Replogle holdout (509 conditions). Sweeping the top-20 amplitude-calibration weight eta shows eta = 0.3 is optimal: larger weights degrade global reconstruction and reduce delta Pearson, confirming that heavy amplitude pressure buys behavioral gains at the expense of expression fidelity.}
\begin{tabular}{llllll}
\toprule
\textbf{Method} & \textbf{Global P.} & \textbf{Delta P.} & \textbf{Top-20} & \textbf{Direction} & \textbf{Recovery} \\
\midrule
DA ($\lambda$=3.0), $\eta$=0 & 0.966 & 0.179 & 0.269 & 0.680 & 1.52 \\
DA+amp, $\eta$=0.3 & 0.932 & 0.192 & 0.318 & 0.722 & 1.78 \\
DA+amp, $\eta$=0.5 & 0.907 & 0.187 & 0.327 & 0.723 & 1.90 \\
DA+amp, $\eta$=1.0 & 0.816 & 0.175 & 0.320 & 0.723 & 1.54 \\
\bottomrule
\end{tabular}
\end{table}

\begin{table}[H]
\centering
\caption{Extended kNN matching sweep. Table S6 shows k = 1, 5, 10, 30, 50. This table extends the sweep to larger k values and additional mixing weights.}
\begin{tabular}{lllll}
\toprule
\textbf{k} & \textbf{alpha} & \textbf{All delta Pearson} & \textbf{Top-20 DE Pearson} & \textbf{Direction} \\
\midrule
50 & 0.5 & 0.189 & 0.271 & 0.555 \\
75 & 0.0 & 0.208 & 0.275 & 0.559 \\
100 & 0.0 & 0.203 & 0.269 & 0.558 \\
\bottomrule
\end{tabular}
\end{table}

\end{document}